\documentclass[final,3p,times,twocolumn]{elsarticle}

\usepackage{amssymb}
\usepackage{amsmath}
\usepackage{multirow}
\usepackage{bm}

\journal{Information Fusion}
\usepackage{booktabs}
\usepackage{longtable}
\usepackage{amsmath,amsfonts}
\usepackage{xcolor}
\usepackage{algorithm}  
\usepackage{algorithmic}  
\usepackage{array}
\usepackage{pifont}
\usepackage{float}

\usepackage{hyperref}
\usepackage[capitalise]{cleveref}
\usepackage{multirow}
\usepackage[table,xcdraw]{xcolor}
\usepackage{graphicx}
\usepackage{booktabs}
\usepackage{enumitem}
\usepackage{colortbl}
\usepackage{adjustbox}
\definecolor{top1}{RGB}{255,0,0} 
\definecolor{top2}{RGB}{0,0,255} 
\definecolor{top3}{RGB}{0,255,0} 

\begin{document}

\begin{frontmatter}



\title{All-weather Multi-Modality Image Fusion: Unified Framework and 100k Benchmark}

\author[author1]{Xilai Li \fnref{co-first}}
\ead{20210300236@stu.fosu.edu.cn}

\author[author1]{Wuyang Liu \fnref{co-first}}
\ead{2112203023@stu.fosu.edu.cn}

\author[author1,label1]{Xiaosong Li\corref{cor}}
\ead{lixiaosong@buaa.edu.cn}

\author[author2]{Fuqiang Zhou}
\ead{zfq@buaa.edu.cn}

\author[author3]{Huafeng Li}
\ead{lhfchina99@kust.edu.cn}

\author[author4]{Feiping Nie}
\ead{feipingnie@gmail.com}

\fntext[co-first]{Equal Contribution}
\cortext[cor]{Corresponding author}


\address[author1]{ Guangdong-HongKong-Macao Joint Laboratory for Intelligent MicroNano Optoelectronic Technology, School of Physics and Optoelectronic Engineering, Foshan University, Foshan 528225, China.}

\address[label1]{Guangdong Provincial Key Laboratory of Industrial Intelligent Inspection Technology, Foshan University, Foshan 528000, China}

\address[author2]{ School of Instrumentation and Optoelectronic Engineering, Beihang University, Beijing 100083, China}

\address[author3]{ School of  Information Engineering and Automation, Kunming University of Science and Technology, Kunming 650500, China}

\address[author4]{ School of Artificial Intelligence, Optics and Electronics (i0PEN), School of Computer Science, Northwestern Polytechnical University, China}

\begin{abstract}
Multi-modality image fusion (MMIF) combines complementary information from different image modalities to provide a comprehensive and objective interpretation of scenes. However, existing fusion methods cannot resist different weather interferences in real-world scenes, limiting their practical applicability. To bridge this gap, we propose an end-to-end, unified all-weather MMIF model. Rather than focusing solely on pixel-level recovery, our method emphasizes maximizing the representation of key scene information through joint feature fusion and restoration. Specifically, we first decompose images into low-rank and sparse components, enabling effective feature separation for enhanced multi-modality perception. During feature recovery, we introduce a physically-aware clear feature prediction module, inferring variations in light transmission via illumination and reflectance. Clear features generated by the network are used to enhance salient information representation. We also construct a large-scale MMIF dataset with 100,000 image pairs comprehensively across rain, haze, and snow conditions, as well as covering various degradation levels and diverse scenes. Experimental results in both real-world and synthetic scenes demonstrate that the proposed method excels in image fusion and downstream tasks such as object detection, semantic segmentation, and depth estimation. The source code is available at
\href{https://github.com/ixilai/AWFusion}{https://github.com/ixilai/AWFusion}.

\end{abstract}


\end{frontmatter}

\section{Introduction}

Infrared and visible image fusion (IVIF) is performed to integrate the complementary and salient feature information from different modal images to achieve a more comprehensive interpretation of a target scene. By integrating data from different spectral ranges, IVIF contributes to advancements in computer vision tasks \cite{r47,r88,r90,r82,r83}, including semantic segmentation \cite{r5,r6}, object detection \cite{r8,r9}, and unmanned driving \cite{r12}. 

Recently, deep learning (DL) has exhibited remarkable progress within the realm of artificial intelligence \cite{r104}, owing to its exceptional aptitude for intricate nonlinear approximations and adeptness in extracting salient features. Consequently, several algorithms tailored to the IVIF have been developed. These algorithms can be broadly categorized into four groups: generative adversarial networks (GAN)-based models \cite{r13,r78}, diffusion-based models \cite{r95,r96}, autoencoder (AE)-based models \cite{r2,r20}, and algorithmic expansion models \cite{r4,r21}. However, most of these methods are designed for ideal imaging conditions. With the increasing need for robust perception in real-world applications, IVIF is expected to deliver reliable scene representation under adverse weather, where visibility is reduced and modality information becomes severely imbalanced.
Although current DL-based fusion methods \cite{r91,r92,r86,r94} achieve promising results in clear-weather scenarios, their generalization and robustness remain limited in rainy, hazy, or snowy environments. Therefore, three major challenges need to be addressed:

\begin{itemize}[label=$\bullet$] 
\item Currently, there is a critical research gap in that IVIF cannot deliver scene information under adverse weather. Addressing this challenge requires more than isolated algorithmic solutions for specific weather-related issues, necessitating a unified framework to comprehensively restore intricate scene details affected by adverse weather.

\item The network design of most current DL-based methods, primarily developed through empirical and experiential approaches, lacks interpretability due to their black-box nature.  As a result, identifying an appropriate framework often becomes a time-consuming process.

\item Generally, DL-based algorithms are driven by data. Existing datasets dedicated to collecting comprehensive imaging data from multiple modalities are complete in terms of scene diversity. However, no large datasets have been developed specifically for research on applying IVIF to scenarios with adverse weather conditions.
\end{itemize}

In summary, it is essential to build a large-scale multi-modality image dataset for specifically tailored for adverse weather scenarios, and to develop an effective unified ``Restoration + Fusion'' framework, enabling more reliable IVIF performance in challenging environments.

\subsection{Our Contributions}
To address above challenges, we propose an all-weather MMIF algorithm (AWFusion) that integrates both deweathering and fusion functions within a single framework, thereby providing an effective means to achieve robust fusion under adverse weather. To clearly present the design principles of the proposed algorithm, we describe it from the following three aspects.

\noindent\textbf{(1) Feasibility Analysis of Unified Framework:}
We argue that to construct a effective unified framework, complementary information should first be integrated without redundancy, followed by restoring the overall clear features. The main reasons are as follow:
(1) In natural scenes, images contain sufficient redundancy, allowing the original content to be inferred even under adverse weather degradation.   From a network learning perspective, deep neural networks extract features at different levels through layer-by-layer feature extraction operations.   Early layers capture low-level cues like edges and textures, while deeper layers encode complex semantics and scene structures.   However, the loss of low-level features does not entirely compromise high-level scene information.   The network can reconstruct the overall scene from remaining valid pixels or local features.   Therefore, it is not necessary to perform image restoration first when handling complex scenes.
(2) The task of IVIF inherently involves discarding redundant information and retaining complementary data.    Unlike image restoration, which aims to recover all details, image fusion focuses on preserving the most significant information.   By leveraging complementary multi-modality features, the network emphasizes useful details while ignoring redundant pixels.   This not only reduces computational complexity but also enhances the ability of the network to recover the aspects of the scene that best represent its true characteristics.  
\textbf{\textit{Thus, our objective is to shift the focus of the network from perfect pixel-level recovery to maximizing the representation of key scene information.}} The unified framework is consisting of two main components: a fusion module and a restoration module.

\noindent\textbf{(2) Implementation of Unified Framework:}
In the fusion module, we introduce a learned low-rank representation (LLRR) block \cite{r4} to decompose the source images into sparse and low-rank components firstly. This feature decoupling approach allows for a more comprehensive consideration of the feature information of the image. Additionally, low-rank and sparse decomposition reduces the computational cost of fusion module: the low-rank component can capture global structure with fewer parameters, while the sparse component focuses on salient pixels, enabling efficient processing of large-scale images. To integrate features,  a Sparse Feature Prediction Block (SFB) was designed to reconstruct sparse components, and a Sparse Transformer Blocks (STB) \cite{r27} was introduced to capture the long-range contextual information of an image. 
In the restoration module, we design a the Physical-Aware Clear Feature Prediction Block (PFB), guided by the atmospheric scattering model \cite{r24}. The PFB derives illumination and reflectance features from images to predict the distribution of light transmission, utilizing learned convolutional sparse coding (LCSC) \cite{r26}. It removes the need for weather-specific model design, enabling simultaneous training of multiple teacher networks in a unified restoration framework. Subsequently, distillation learning transfers knowledge from multiple teacher networks to a single student network, enabling the model to handle adverse weather degradations.

\noindent\textbf{(3) Construction of Large-Scale All-Weather Benchmark:}
Furthermore, we introduce a large-scale benchmark, AWMM-100k, specifically designed for multi-modality image fusion in adverse weather conditions, aiming to advance research in this challenging field. Our dataset consists of 100,000 multi-modality image pairs, comprehensively covering both synthetic and real-world scenes affected by three major weather degradations: rain, haze, and snow. To ensure robust evaluation, the data spans multiple degradation levels for each weather type, capturing the diversity and complexity of natural environments. This benchmark addresses a critical gap in the field by providing a diverse, and extensive resource for developing and evaluating fusion methods under real-world adverse weather.

The significant contributions of this work are summarized below:

\begin{itemize}[label=$\bullet$] 
\item We propose an end-to-end unified all-weather MMIF framework. To the best of our knowledge, this is the first work to address MMIF and adverse weather removal simultaneously, defining a new challenging task.

\item We propose a feature separation-based fusion module that considers the reconstruction of local and global features from sparse and low-rank aspects, and develop a physically-aware feature prediction block that comprehensively accounts for the restoration of clear features.

\item We construct an all-weather MMIF benchmark (AWMM-100k) contains rain, haze, and snow scenes with different level. AWMM-100k covers real-world scenes, and has totally 100,000 registered multi-modality image pairs with object detection labels.

\item Extensive experiments on adverse weather image fusion, as well as downstream tasks cover depth estimation, semantic segmentation, and object detection, verify the effectiveness and superiority of the proposed method.
\end{itemize}
\begin{figure*}[t]
  \centering
   \includegraphics[width=1.0\linewidth]{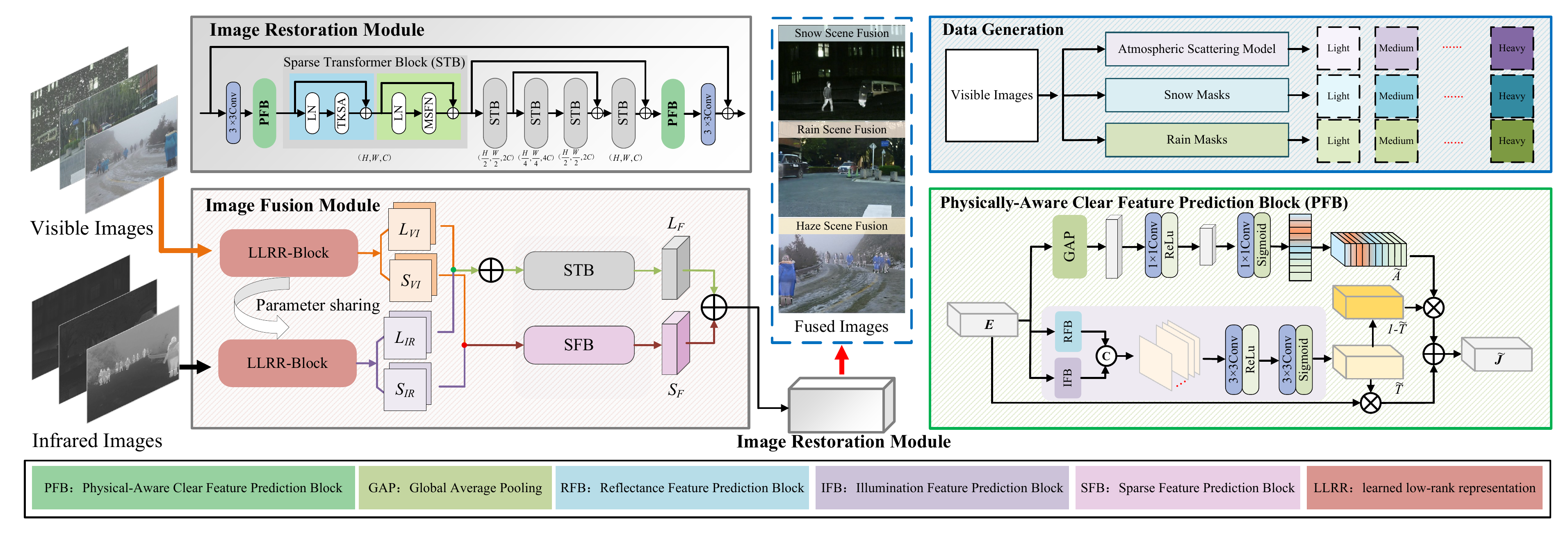}
   \caption{The architecture of the proposed all-weather MMIF framework. The proposed end-to-end method comprises an image fusion module and an image restoration module. Additionally, a rough outline of the data generation process is illustrated in the upper right corner.}
   \label{fig1}
\end{figure*}

\section{Related Work}
\subsection{Image Fusion in Conventional Conditions}

DL-based IVIF algorithms \cite{r31,r79,r75,r84,r73,r74,r76,r77} demonstrate robust fusion performance in conventional scenes, employing methodologies such as GAN or AE to effectively integrate salient feature information from each modality. 
For example, Zhao et al. \cite{r66} introduced a self-supervised framework incorporating equivariant priors to address the lack of ground truth in multimodal image fusion.  Similarly, Cheng et al. \cite{r86} exploited underlying visual tasks in digital photography fusion as supervisory signals and proposed a generalized fusion network that enhances cross-task interaction.
With the widespread utilization of IVIF technology, researchers increasingly integrate this technology into downstream vision tasks \cite{r85,r33,r34,r54}. Their aim is to enhance the extraction of multi-modality features, augmenting semantic information and leveraging fused images to optimize the performance of tasks such as object detection, semantic segmentation, and salient object detection. 
For instance, Tang et al. \cite{r34} argued that most image fusion methods overlook the demands of high-level visual tasks, leading to a lack of semantic information in the fused results. To address this, they proposed a semantic-aware real-time image fusion network. Liu et al. \cite{r87} noted that existing approaches often treat fusion and downstream tasks as separate processes, which limits the benefits of fusion for subsequent tasks. To overcome this, they introduced a discriminative cross-dimensional evolutionary learning framework for IVIF, employing an evolutionary algorithm to optimize the loss function parameters.

\subsection{Image Fusion in Complex Conditions}

Additionally, there are many joint multi-tasking algorithms designed to cope with real-world applications \cite{r71,r36,r80}.
In practical applications, images captured by cameras may be slightly misaligned and distorted; some researchers \cite{r35} have designed correction networks for fusion specifically for such situations. Xu et al. \cite{r36} designed an IVIF model that could fuse unaligned image pairs, effectively reducing the cost of adopting IVIF technology. In some cases, fixed input and output resolutions constrain the development of IVIF methods. To address this limitation, Li et al. \cite{r37} To address this limitation, Li et al. proposed a meta-learning-based IVIF framework, where a meta-amplification module dynamically predicts amplification filter weights using scale factors as input.
In addition, some algorithms apply the IVIF task to degraded environments, such as noise \cite{r68}, low light \cite{r69}, and low contrast scenes \cite{r67}. For example, Li et al. \cite{r97} proposed a unified IVIF framework for complex scenes based on optical-transmission decomposition.  The framework leverages the prior that degradation factors cause scattering during light transmission, allowing the transmission map to capture the distribution of different degradations. By separating the degraded features from the source image, the loss of detail in the feature extraction process is avoided. Zhang et al. \cite{r98} combined multi-modal large models to develop a robust, language-driven image fusion algorithm.  It leverages the denoising mechanism of latent diffusion models to progressively remove degraded information and can locate targets of interest based on language instructions, highlighting specific scene details. However, current research does not effectively address the generalization of IVIF algorithms, whether they are single-task or multi-task, to adverse weather conditions.

\section{Proposed Method}

\subsection{Problem Formulation}

Given an infrared and visible image pair $\{{I}_{{IR}}, {I}_{{VI}}\}$ captured under adverse weather, our objective is to generate a fused image ${I}_{F}$ that retains complementary modality information while recovering a clean visual scene. The task can be formulated as:
\begin{equation}
{I}_{F} = \mathcal{F}({I}_{{IR}}, {I}_{{VI}}; \Theta),
\end{equation}
where $\mathcal{F}(\cdot)$ denotes the proposed AWFusion network and $\Theta$ represents the learnable parameters. This formulation involves two key goals: (1) effective cross-modal feature fusion and (2) restoration of weather-degraded structures and textures.

\noindent\textbf{Fusion module.}
To disentangle complementary information, the input images are first decomposed into low-rank and sparse components using the LLRR block:
\begin{equation}
\begin{aligned}
({L}_{{VI}}, {S}_{{VI}}) &= \mathrm{LLRR}\left({I}_{{VI}}\right), \\
({L}_{{IR}}, {S}_{{IR}}) &= \mathrm{LLRR}\left({I}_{{IR}}\right),
\end{aligned}
\end{equation}
where $\mathrm{LLRR}(\cdot)$ denotes the processing performed by the Learned Low-Rank Representation Blocks, 
${L}_{{VI}}, {L}_{{IR}}$ represent the global structure (low-rank components), 
and ${S}_{{VI}}, {S}_{{IR}}$ contain modality-specific salient details (sparse components).

The two components are then fused separately to better preserve shared structures and unique characteristics:
\begin{equation}
\begin{aligned}
{L}_{{F}} &= \mathrm{STB}\big({L}_{{VI}}, {L}_{{IR}}), \\
{S}_{\text{F}} &= \mathrm{SFB}\big({S}_{{VI}}, {S}_{{VI}}),
\end{aligned}
\end{equation}
Where $\mathrm{STB}(\cdot)$ and $\mathrm{SFB}(\cdot)$ denote the processing performed by the Sparse Transformer Blocks and the Sparse Feature Prediction Blocks, respectively, producing low-rank fused features ${L}_{{F}}$ and sparse fused features ${S}_{{F}}$, which are then aggregated and fed into the restoration module for clear-scene reconstruction.

\noindent\textbf{Restoration module.} The restoration module employs a U-Net backbone augmented with five STB, each consisting of a Top-$k$ Sparse Attention (TKSA) mechanism and a Mixed-Scale Feedforward Network (MSFN) \cite{r27}. TKSA aggregates the most informative features while maintaining sparsity, and MSFN extracts multi-scale local information. Additionally, the PFB refines spatial information and is embedded on both sides of the backbone to improve feature representation. The restored fusion result is then output as the final prediction $I_F$.

Finally, to enhance the robustness of model under adverse weather conditions, we introduce a distillation learning strategy \cite{r38}. Specifically, multiple teacher networks are first trained for distinct weather degradations, and their knowledge is then transferred to a single student network to comprehensively address adverse weather effects.

\subsection{Image Fusion Module}
Due to significant differences between modalities and potential semantic conflicts in certain cases, directly fusing them in the spatial domain can lead to insufficient feature interaction, contrast loss caused by modality misalignment, or misalignment artifacts. Therefore, decomposing image features prior to fusion facilitates the network in capturing interactions among different types of features, enabling a better understanding of the semantic information.
Therefore, we decompose the image into low-rank and sparse components through the LLRR blocks. Low-rank components reflect the global information and main structure of the image, constrained by certain rank features. Given the strong ability of Transformer to capture global context, we employ the STB to fuse low-rank features from different modalities. In contrast, sparse components represent local details such as edges, textures, and high-frequency information. To effectively manage these local fusions, we design the SFB.
\subsubsection{Sparse Feature Prediction Block (SFB)}
Sparse feature fusion is performed to effectively extract and integrate local features specific to infrared images ($IR$) and visible images ($Vis$). Here, we employ an optimization principle inspired by LCSC to guarantee that the learned sparse filters represent meaningful high-frequency image details. This is motivated by the physical observation that edges and textures dominate local image variations, and proper modeling is necessary to preserve them in the fused output. 

The fused feature map ${S}_{{F}}$ consists of two parts: the sparse feature $s_{I,k} \in \mathbb{R}^{H \times W \times K}$ for $IR$ and the sparse feature $s_{V,k} \in \mathbb{R}^{H \times W \times K}$ for $Vis$, modeled as:
\begin{equation}
  {S}_{{F}} = \sum_{k} g_k^{S_I} \ast s_{I,k} + \sum_{k} g_k^{S_V} \ast s_{V,k}
  \label{eq13}
\end{equation}
where $\left\{ g_k^{S_I} \right\}_{k=1}^K$ and $\left\{ g_k^{S_V} \right\}_{k=1}^K$ are the convolutional filters for $IR$ and $Vis$, respectively.

To derive a tractable optimization problem from the convolutional representation in \cref{eq13}, we first express the convolutions in matrix-vector form. Each filter $g_k^{S_I}$ and $g_k^{S_V}$ is converted into a Toeplitz matrix $G_k^{S_I}$ and $G_k^{S_V}$, while the corresponding feature maps $s_{I,k}$ and $s_{V,k}$ are vectorized. This allows the fused feature to be equivalently written as a linear combination of sparse components:
\begin{equation}
    {S}_{{F}} = \sum_k G_k^{S_I} s_{I,k} + \sum_k G_k^{S_V} s_{V,k}
\end{equation}
Next, to ensure that the sparse features accurately reconstruct the original images, we introduce reconstruction fidelity terms for each modality, measured by the $\ell_2$-norm of the difference between the source images and their reconstructions:
\begin{equation}
    \frac{1}{2}\Big\| IR - \sum_k G_k^{S_I} s_{I,k} \Big\|_2^2 + \frac{1}{2}\Big\| Vis - \sum_k G_k^{S_V} s_{V,k} \Big\|_2^2
\end{equation}
Finally, to encourage sparsity in the learned features and suppress redundant activations, we add an $\ell_1$ regularization term for each sparse feature. Combining the fidelity and sparsity terms naturally leads to a joint optimization problem that can be solved using convolutional sparse coding.
The sparse feature responses $\left\{ s_{I,k},s_{V,k} \right\}_{k=1}^K$ are then obtained by solving:
\begin{equation}
\begin{aligned}
\mathrm{Argmin}_{{S}_k^I,{S}_k^V} & \frac{1}{2} \left\| IR - \sum_{k} (g_k^{S_I} \ast s_{I,k}) \right\|_2^2 \\ + \frac{1}{2} \left\| Vis - \sum_{k} (g_k^{S_V} \ast s_{V,k}) \right\|_2^2 & + \lambda \sum_{k} \left( \|s_{I,k}\|_1 + \|s_{V,k}\|_1 \right)
\end{aligned}
\end{equation}
Similar to \cite{r21}, $s_{I,k}$ and $s_{V,k}$ are solved as follows:
\begin{equation}
  S_{j+1}^I = S_{\lambda} (S_j^I - L_{S_I} \ast C_{S_I} \ast S_j^I + L_{S_I} \ast IR)
  \label{eq11}
\end{equation}
\begin{equation}
  S_{j+1}^V = S_{\lambda} (S_j^V - L_{S_V} \ast C_{S_V} \ast S_j^V + L_{S_V} \ast Vis)
  \label{eq12}
\end{equation}
where $S_I \in \mathbb{R}^{H \times W \times K}$ and $S_V \in \mathbb{R}^{H \times W \times K}$ are the stack of feature responses $\left\{ s_{I,k} \right\}_{k=1}^K$ and $\left\{ s_{V,k} \right\}_{k=1}^K$ respectively; $S_j^I$ and $S_j^V$ are the updates of $S_I$ and $S_V$ at the $j$-th iteration, respectively; and $C_{S_I}\in \mathbb{R}^{s \times s \times K}$, $C_{S_V}\in \mathbb{R}^{s \times s \times K}$, $L_{S_I}\in \mathbb{R}^{s \times s \times K}$ and $L_{S_V}\in \mathbb{R}^{s \times s \times K}$ are the learnable convolutional layers.
According to LCSC, we can expand the iterations in \cref{eq11} and \cref{eq12} into a network as shown in the framework in \cref{fig2}.

\subsection{Image Restoration Module}
The second phase aims to achieve image restoration in the presence of three types of adverse weather conditions (rain, snow, and haze). 
Based on the Atmospheric Scattering Model (ASM), adverse weather conditions such as rain, haze, and snow induce scattering and absorption of light by particles in the atmosphere. Consequently, these atmospheric phenomena lead to comparable image degradation across a spectrum of unfavorable conditions \cite{r23,r97}. For instance, during heavy rainfall, the accumulation of distant rain lines and atmospheric water particles creates a "rain curtain" effect. This visual effect of rainwater accumulation produces a haze like phenomenon in the image background \cite{r48}. The above conclusion provide a research basis to realize deweathering in the same framework. 

To leverage this unified representation, we construct the PFB guided by the ASM. Specifically, the network is designed to predict the key parameters of the ASM, including the atmospheric light and the transmission map, using learnable neural modules. By estimating these parameters, the network can effectively model the scattering and absorption effects induced by adverse weather. Subsequently, the clear image features are reconstructed by combining the predicted atmospheric light and transmission map, enabling the recovery of scene information in a physically meaningful and interpretable manner. This approach enables the network to learn from data while incorporating physical degradation priors.

\subsubsection{Physical-Aware Clear Feature Prediction Block (PFB)}
The construction of the PFB was guided by the physical mechanism of image degradation and transformation of the atmospheric scattering model. First, utilizing kernel \textit{k} for feature extraction results in the following revised expression of atmospheric scattering model:

\begin{equation}
  k \ast J = k \ast (I \odot \frac{1}{T}) + k \ast A - k \ast (A \odot \frac{1}{T})
  \label{eq2}
\end{equation}
\noindent where $\ast$ denotes the convolution operator, and $\odot$ represents the Hadamard product. Subsequently, by introducing the matrix-vector forms of $k$, $J$, $I$, $A$, and $1/T$, which are denoted as ${K}$, ${J}$, ${I}$, ${A}$, and ${D}$, respectively, \cref{eq2} can be redefined as follows:
\begin{equation}
{KJ} = {KDI} + {KA} - {KDA}
  \label{eq3}
\end{equation}
\noindent According to \cite{r24}, given the input adverse weather image feature ${E}\in\mathbb{R}^{H \times W \times K}$, the clear feature $\tilde{{J}}\in\mathbb{R}^{H \times W \times K}$ can be computed as follows:
\begin{equation}
  \tilde{{J}} = {E} \odot \tilde{{T}} + \tilde{{A}} (1 - \tilde{{T}})
  \label{eq4}
\end{equation}
\noindent where $\tilde{{A}}\in\mathbb{R}^{H \times W \times K}$ represents an approximation of the characteristic $\textbf{\textit{KA}}$ of the atmospheric light $A$, and $\tilde{{T}}\in\mathbb{R}^{H \times W \times K}$ is a characteristic associated with the transmission map. 1 represents a matrix composed entirely of ones. Since atmospheric light is usually assumed to be homogeneous, as shown in \cref{fig1}, we use global average pooling to eliminate redundant information within the feature space to predict the atmospheric light feature $\tilde{{A}}$.

In adverse weather conditions like dense haze, rain, or snow, light undergoes scattering and absorption, causing alterations in both image reflectance and illumination. For example, haze reduces the intensity of light reflected from objects to the camera by scattering it, with this effect being more pronounced at greater distances. Rain and snow also alter the surface reflectivity of objects when they adhere to them, affecting how light is reflected. Additionally, atmospheric particles can cause light to be reflected, refracted, and absorbed, impacting the overall illumination of the scene. Consequently, for the prediction of the feature $\tilde{{T}}$, by analyzing changes in reflectance and illumination of objects in an image, we can infer changes in light transmission, thus realizing estimation of the transmission information. 

We disassemble $\tilde{{T}}$ into reflectance $d_k\in\mathbb{R}^{H \times W \times K}$ and illumination $m_k\in\mathbb{R}^{H \times W \times K}$ components, modeled as:
\begin{equation}
  \tilde{{T}} = \sum_{k} g_k^d \ast d_k + \sum_{k} g_k^m \ast m_k
  \label{eq5}
\end{equation}
where $\left\{ g_k^d \right\}_{k=1}^K$ and $\left\{ g_k^m \right\}_{k=1}^K$ are the convolutional filters for reflectance and illumination, respectively.

To derive a tractable optimization formulation from the convolutional decomposition in \cref{eq5}, we first express the convolutions in matrix-vector form. Each filter $g_k^d$ and $g_k^m$ is converted into a Toeplitz matrix $G_k^d$ and $G_k^m$, while the corresponding feature maps $d_k$ and $m_k$ are vectorized. This allows the decomposition to be equivalently written as a linear combination of reflectance and illumination components:
\begin{equation}
    \tilde{{T}} = \sum_k G_k^d d_k + \sum_k G_k^m m_k
\end{equation}
Then, we introduce reconstruction fidelity terms for both components, measured by the $\ell_2$-norm of the difference between $I$ and their reconstructions:
\begin{equation}
    \frac{1}{2}\Big\| I - \sum_k G_k^d d_k \Big\|_2^2 + \frac{1}{2}\Big\| I - \sum_k G_k^m m_k \Big\|_2^2
\end{equation}
Lastly, to encourage sparsity in the learned reflectance and illumination features and suppress redundant activations, we add an $\ell_1$ regularization term for each feature map. Combining the fidelity and sparsity terms leads to a joint optimization problem that can be solved using convolutional sparse coding.

Accordingly, the reflectance and illumination feature responses $\left\{ d_k,m_k \right\}_{k=1}^K$ are obtained by solving:
\begin{equation}
\begin{aligned}
\mathrm{Argmin}_{{d}_k,{m}_k} & \frac{1}{2} \left\| I - \sum_{k} (g_k^d \ast d_k) \right\|_2^2 \\ + \frac{1}{2} \left\| I - \sum_{k} (g_k^m \ast m_k) \right\|_2^2 & + \lambda \sum_{k} \left( \|d_k\|_1 + \|m_k\|_1 \right)
\end{aligned}
\label{eq16}
\end{equation}
Furthermore, We emphasize that our objective is adverse weather restoration rather than strict Retinex-based intrinsic decomposition. In our framework, reflectance $d_k$ and illumination $m_k$ serve as two feature subspaces for estimating transmission-related cues in the atmospheric scattering model. Although a shared sparsity prior is applied, convolutional filters and loss fusion still guide the model to learn distinct responses, effectively separating scene details from low-frequency degradation.


\begin{figure}[t]
  \centering
   \includegraphics[width=1.0\linewidth]{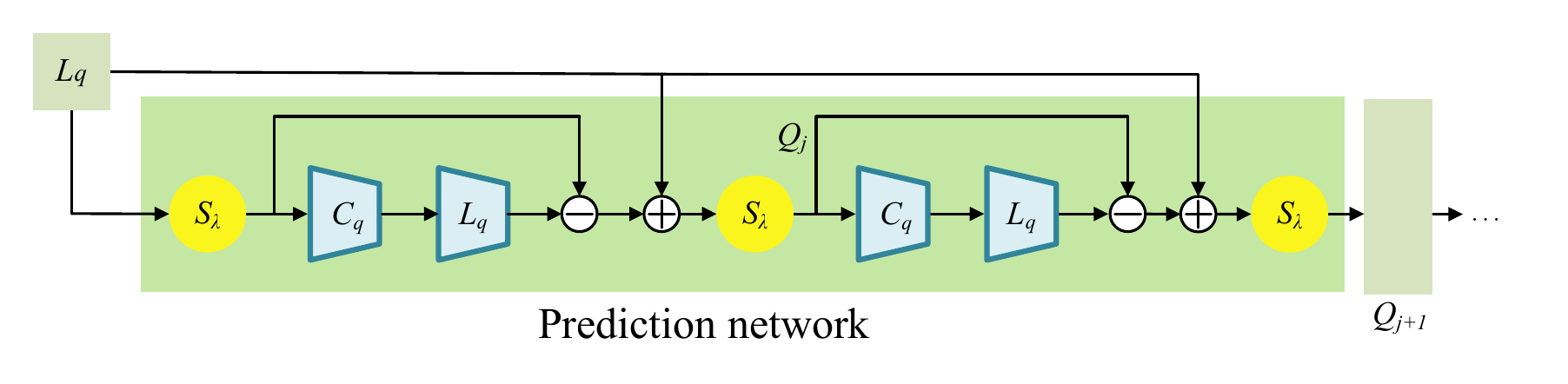}
   \caption{Network structure of RFB, IFB, and SFB. $q_k \in \{d_k, m_k, s_{I,k}, s_{V,k}\}$ and $Q_j \in \{D_j, M_j, S_j^I, S_j^V\}$.}
   \label{fig2}
\end{figure}
\paragraph{\textbf{Reflectance Feature Prediction Block (RFB)}}
The RFB is designed to extract reflectance features, we formalize this as a convolutional sparse coding problem. According to \cite{r21}, we fix $m_k$ to update $d_k$ by rewriting \cref{eq16} as: 
\begin{equation}
  \underset{d_k}{\mathrm{Argmin}} \frac{1}{2} \left\| I - \sum_{k} (g_k^d \ast d_k) \right\|_2^2 + \lambda \sum_{k} \|d_k\|_1
  \label{eq7}
\end{equation}
\noindent \cref{eq7} is a standard convolutional sparse coding problem that can be solved using LCSC \cite{r26} by rewriting \cref{eq7} as:
\begin{equation}
  D_{j+1} = S_{\lambda} \left( D_j - L_d \ast C_d \ast D_j + L_d \ast I \right)
  \label{eq8}
\end{equation}
\noindent where $D \in \mathbb{R}^{H \times W \times K}$ is the stack of feature responses  $\left\{ d^k \right\}_{k=1}^K$, $D_j$ is the update of $D$ at the $j$-th iteration, $C_d \in \mathbb{R}^{s \times s \times K}$ and $L_d \in \mathbb{R}^{s \times s \times K}$ are the learnable convolutional layers related to the filters $\left\{ g_k^d \right\}_{k=1}^K$, and $S_{\lambda}$ represents the soft thresholding operator. Similarly, we expand the iterations in \cref{eq8} into a network as shown in \cref{fig2}.

\paragraph{\textbf{Illumination Feature Prediction Block (IFB)}}
The goal of IFB is to compute the illuminance feature in the degraded image, the structure of this network is similar to that of RFB. By fixing $d_k$ to update $m_k$, \cref{eq16} can be rewritten as: 
\begin{equation}
  \text{Argmin}_{m_k} \frac{1}{2} \|I - \sum_{k} g_k^d \ast m_k \|_2^2 + \lambda \sum_{k} \|m_k \|_1
  \label{eq9}
\end{equation}
\noindent By using LCSC to solve the above problem, \cref{eq9} can be rewritten as follows:
\begin{equation}
  M_{j+1} = S_{\lambda} (M_j - L_m \ast C_m \ast M_j + L_m \ast I)
  \label{eq10}
\end{equation}
\noindent where $M \in \mathbb{R}^{H \times W \times K}$ is the stack of feature responses $\left\{ m^k \right\}_{k=1}^K$, $M_j$ is the update of $M$ at the $j$-th iteration, and $C_m \in \mathbb{R}^{s \times s \times K}$ and $L_m \in \mathbb{R}^{s \times s \times K}$ are the learnable convolutional layers related to the filters $\left\{ g_k^m \right\}_{k=1}^K$. The network structure of the IFB is shown in \cref{fig2}.

\subsection{Loss Function}
The loss function consists of two main components: the first corresponds to the training loss of the teacher networks, while the second pertains to the training loss of the student network.  The former is primarily based on standard image fusion losses, whereas the latter incorporates knowledge distillation.

\subsubsection{Teacher Networks Loss Function}
This loss function integrates two components. The first part focuses on the loss function used to ensure the quality of the fusion results, while the second part addresses the loss function constraining the generation of reflectance and illumination features. 
The fusion loss includes structural similarity index (SSIM) \cite{r40}, mean square error (MSE) and color consistency loss. The calculation for SSIM can be expressed as follows:
\begin{equation}
\text{SSIM}\left(X,\ Y\right)=\frac{\left(2\mu_X\mu_Y+c_1\right)\left(2\sigma_{XY}+c_2\right)}{\left(\mu_X^2+\mu_Y^2+c_1\right)\left(\sigma_X^2+\sigma_Y^2+c_2\right)}
\label{eq14}
\end{equation}
where $X$ and $Y$ represents two different images. The $\mu_Z$, where $Z\in {X, Y}$, denotes the mean of image $Z$, $\sigma_Z$ represents the standard deviation of image $Z$, $\sigma_{XY}$ denotes the covariance between images $X$ and $Y$. The constants $c_1$ and $c_2$ are introduced to prevent the denominator from approaching zero. Thus, the formula for the SSIM loss ($\mathcal{L}_{\text{SSIM}}$) is as follow:
\begin{equation}
\begin{split}
\mathcal{L}_{\text{SSIM}}\left(F, GT_{IR}, GT_{Vis}\right) &= \text{SSIM}\left(GT_{IR}, F\right) \\
&+ \text{SSIM}\left(GT_{Vis}, F\right)
\end{split}
\label{eq15}
\end{equation}
where $GT_{IR}$ and $GT_{Vis}$ represent the ground truth of infrared and visible images under adverse weather, respectively. In addition, MSE can be calculated as follows:
\begin{equation}
\text{MSE}\left(X,\ Y\right)=\frac{1}{HW}\sum_{i=1}^{H}\sum_{j=1}^{W}{{\left[X\left(i,j\right)-Y\left(i,j\right)\right]\ }^2}
\label{eq_MSE}
\end{equation}
 where $H$ and $W$ represent the height and width of the image, respectively. The MSE loss ($\mathcal{L}_{\text{MSE}}$) is as follow:
\begin{equation}
\begin{split}
\mathcal{L}_{\text{MSE}}\left(F, GT_{IR}, GT_{Vis}\right) &= \text{MSE}\left(GT_{IR}, F\right) \\
&+ \text{MSE}\left(GT_{Vis}, F\right)
\end{split}
\label{eq17}
\end{equation}

Colour consistency loss \cite{r67} is used to ensure that the fused image maintains consistent colours with the $Vis$. We transform the image into the YCbCr color space and apply a constraint using the Euclidean distance between the Cb and Cr channels. This can be expressed as:
\begin{equation}
\begin{split}
\mathcal{L}_{\text{color}}\left(F, GT_{Vis}\right) &= \frac{1}{HW} \left\lVert \mathcal{F}_{\text{CbCr}}(F) - \mathcal{F}_{\text{CbCr}}(GT_{\text{Vis}}) \right\rVert_1
\end{split}
\label{eq18}
\end{equation}
where $\mathcal{F}_{\text{CbCr}}$ represents the transfer function from RGB to the CbCr color space. 

To effectively constrain the generation of the reflectance feature $d_{k}$ and the illuminance feature $m_{k}$ in \cref{eq5}, we propose a reconstruction loss inspired by retina theory. According to this theory, a natural image can be viewed as the product of its reflectance and luminance components. Based on this principle, we reconstruct the natural image by multiplying the extracted reflectance and illuminance features, and impose the reconstruction loss using the $GT_{Vis}$ as the reference. The reconstruction loss can be expressed as follows:
\begin{equation}
  \mathcal{L}_{\text{recon}}\left(d_k, m_k, GT_{Vis}\right) = \left\lVert GT_{Vis} - (d_k \ast m_k) \right\rVert_1
  \label{eq19}
\end{equation}
The total loss of teacher network is denoted as:
\begin{equation}
\begin{split}
  \mathcal{L}_{\text{teacher}} = &\ \mathcal{L}_{\text{SSIM}}\left(F, GT_{IR}, GT_{Vis}\right) 
  + \mathcal{L}_{\text{MSE}}\left(F, GT_{IR}, GT_{Vis}\right) \\
  &+ \mathcal{L}_{\text{color}}\left(F, GT_{Vis}\right) 
  + \mathcal{L}_{\text{recon}}\left(d_k, m_k, GT_{Vis}\right)
\end{split}
\label{eq20}
\end{equation}

\subsubsection{Student Networks Loss Function}
The student network learns from teacher networks trained for three types of adverse weather via knowledge distillation.  Its training objective builds on the teacher network loss while incorporating a Collaborative Knowledge Transfer loss ($\mathcal{L}_{\text{CKT}}$) and a Soft Contrastive Regularization loss ($\mathcal{L}_{\text{SCR}}$)\cite{r38}. 

The $\mathcal{L}_{\text{CKT}}$ consists of the Projection Feature Error loss ($\mathcal{L}_{\text{PFE}}$) and Projection Feature Validation losses $\mathcal{L}_{\text{PFV}}$.  $\mathcal{L}_{\text{PFE}}$ projects teacher and student feature maps into a shared lower-dimensional space to mitigate the impact of architectural differences on knowledge transfer, while $\mathcal{L}_{\text{PFV}}$ ensures that the feature error computed in $\mathcal{L}_{\text{PFE}}$ is based on informative and valid representations.  In essence, this loss does not treat teacher features as hard targets;  instead, the projection and validation mechanisms address feature space misalignment across networks.  
The $\mathcal{L}_{\text{SCR}}$ treats outputs from well-performing teacher networks as positive (soft) targets and the original degraded images as negative samples, enabling the student network to learn effective weather-specific restoration solutions and facilitating smoother knowledge integration during early training.

In summary, the total loss of student network is denoted as:
\begin{equation}
\begin{split}
  \mathcal{L}_{\text{student}} = &\mathcal{L}_{\text{teacher}} 
  + \mathcal{L}_{\text{CKT}} +\mathcal{L}_{\text{SCR}}
\end{split}
\end{equation}

\begin{table}[t]
\caption{Overview of the AWMM-100k dataset: A large-scale multi-modality benchmark for adverse weather conditions.}
\resizebox{\columnwidth}{!}{%
\centering
\begin{tabular}{cl|cl}
\hline
\multicolumn{2}{c|}{Aspect}             & \multicolumn{2}{c}{Description}                           \\ \hline
\multicolumn{2}{c|}{Total Image Pairs}  & \multicolumn{2}{c}{100,000 pairs}                         \\
\multicolumn{2}{c|}{Modalities}         & \multicolumn{2}{c}{Visible (RGB), Infrared (IR)}          \\
\multicolumn{2}{c|}{Weather Conditions} & \multicolumn{2}{c}{Rain, Haze, Snow}                      \\
\multicolumn{2}{c|}{Degradation Levels} & \multicolumn{2}{c}{Heavy, Medium, and Light} \\
\multicolumn{2}{c|}{Data Types}         & \multicolumn{2}{c}{Synthetic and Real-world Scenes}       \\
\multicolumn{2}{c|}{Scenes}             & \multicolumn{2}{c}{Urban, Suburban, Rural, Natural, etc.} \\
\multicolumn{2}{c|}{Task Labels}        & \multicolumn{2}{c}{Object Detection}                      \\ \hline
\end{tabular}
}
\label{AWMM}
\end{table}

\begin{figure*}[t]
  \centering
   \includegraphics[width=1\linewidth]{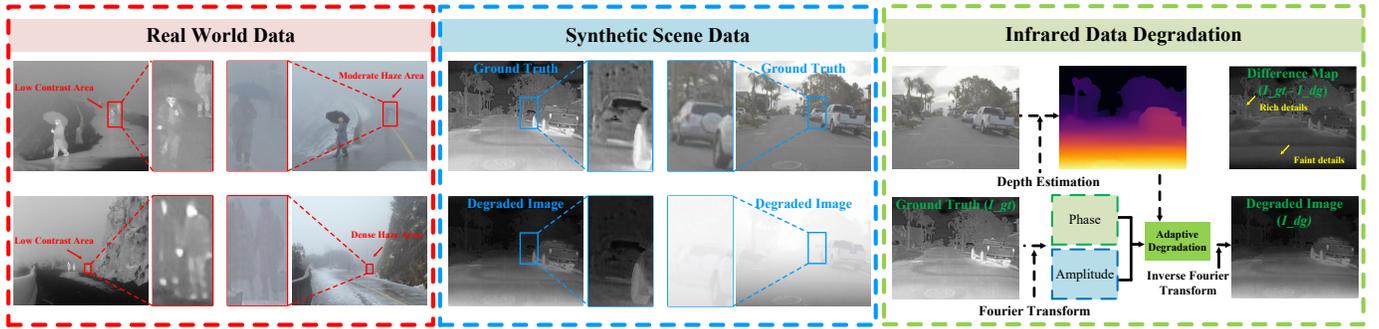}
   \caption{Schematic representation of real-world and synthetic scene data. In this case, the data for the real scene is from AWMM-100k and the data for the synthesized scene is from the RoadScene dataset.}
   \label{fig3}
\end{figure*}

\subsection{MMIF Benchmark for Adverse Weather}
\paragraph{\textbf{Dataset Overview}}  
Current MMIF datasets lack comprehensive scene coverage under adverse weather. Combining existing datasets (RoadScene\cite{r32}, MSRS\cite{r41}, M3FD\cite{r3}, and LLVIP\cite{r42}) and our newly collected data in real-world scenes, we have established a benchmark AWMM-100k, which encompasses a diverse range of weather conditions. Each type of weather (rain, haze, and snow) is categorized into heavy, medium, and light levels. Specific descriptions are shown in \cref{AWMM}. The AWMM-100k dataset was collected using a DJI M30T drone, equipped with  visible and thermal imaging cameras.

\paragraph{\textbf{Challenges in Data Collection}}  
Collecting a large amount of aligned paired multi-modality data and corresponding ground truth in real adverse weather is challenging, leading most current research to rely on synthetic data for training. Effectively degrading clean data and imbuing the model with adaptability to real scenes are pivotal for constructing benchmark. However, the degradation of infrared images in severe weather remains a current research gap. To address this, we analyze data collected during real severe weather to determine the most effective methods for degrading infrared images. Observations of the data in \cref{fig3} allow us to draw the following conclusions: 
\begin{itemize}[label=$\bullet$] 
\item Visible images show clear weather-related degradation, while infrared images exhibit only general contrast and detail reduction without explicit weather cues.  

\item The depth of the scene affects the extent of light refraction and scattering, resulting in greater blurring in distant scenes compared to closer ones. 
\end{itemize}
Building on these observations, we proposed a adaptive degradation strategy for multi-modality images.

\begin{figure}[t]
  \centering
   \includegraphics[width=1\linewidth]{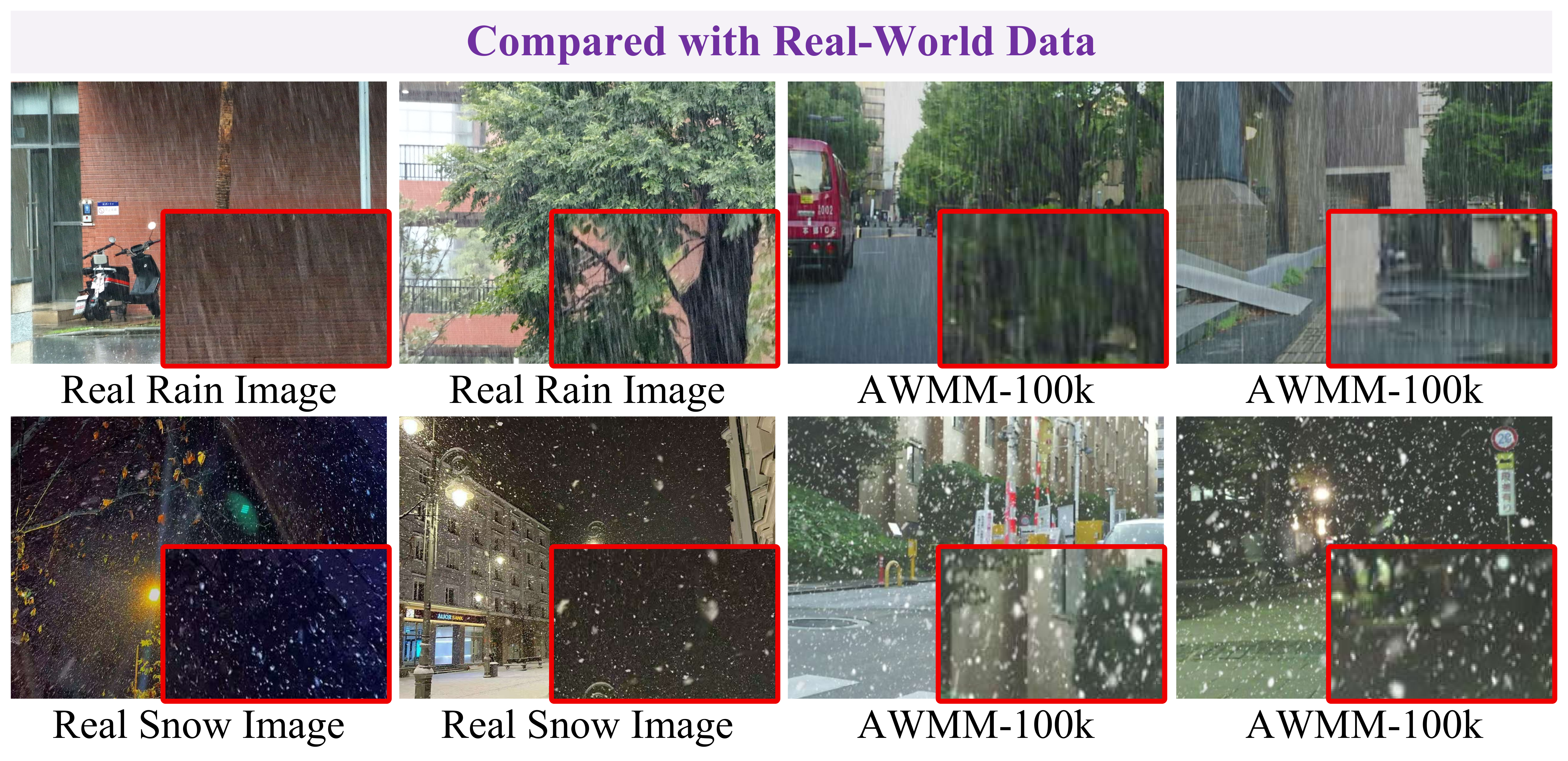}
   \caption{Comparison between synthetic data in AWMM-100k and real degraded data.}
   \label{fig_real_comp}
\end{figure}
\begin{figure}[t]
  \centering
   \includegraphics[width=1\linewidth]{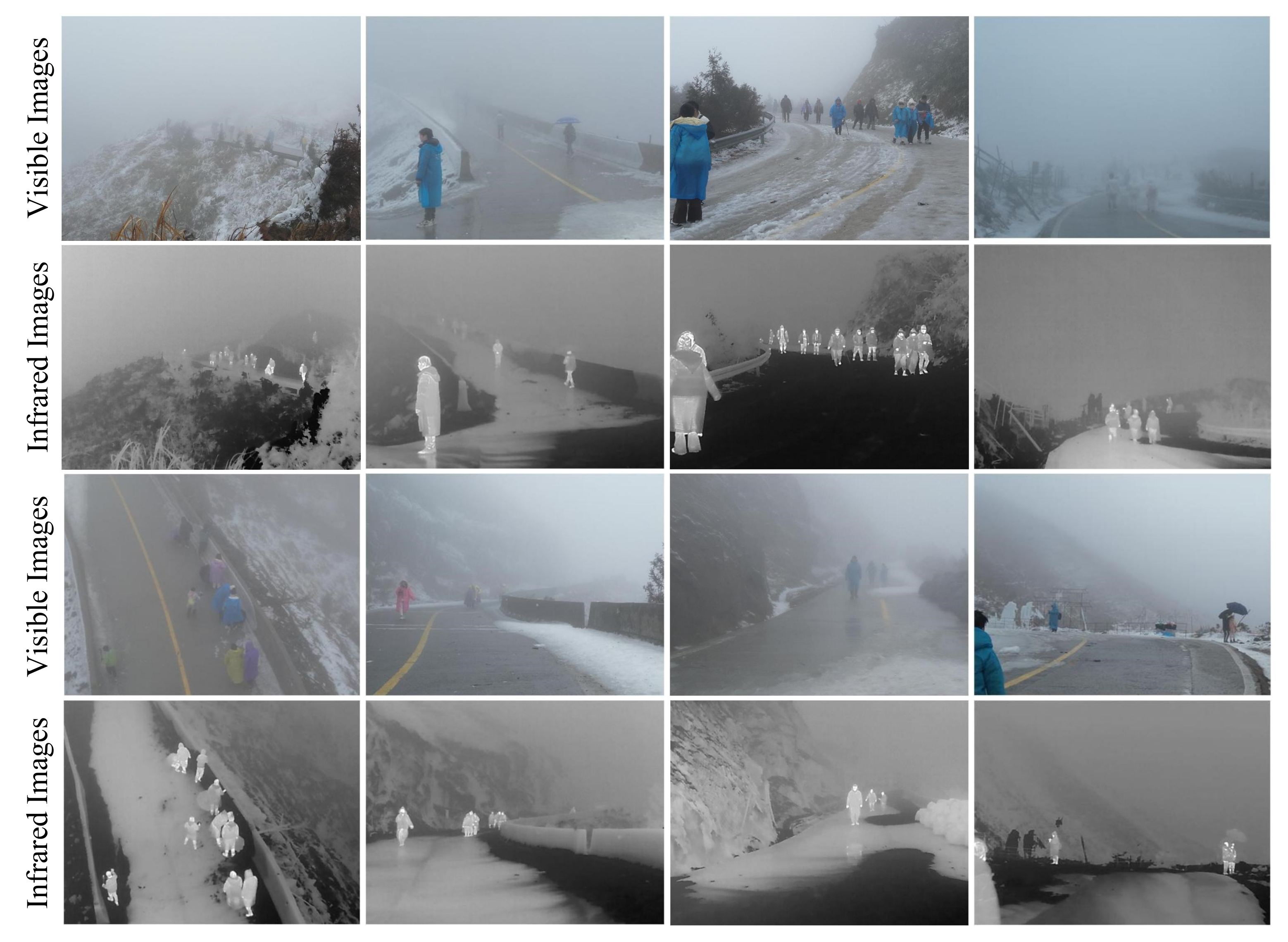}
   \caption{Partial real multi-modality data from AWMM-100k.}
   \label{fig4}
\end{figure}

\paragraph{\textbf{Infrared Degradation Strategy}}  
First, we transpose the image into the Fourier domain and use the depth map as the weights for degradation. Subsequently, the weight map is normalised and multiplied with the amplitude and phase of the image respectively and then inverted to obtain the degraded image. However, distant pixels in the depth map are always represented in black, i.e., the pixel value is $0$. In order to suppress too small weights, we add $0.3$, $0.2$ and $0.1$ to points with pixel values less than $0.2$ according to the three degrees of light, medium and heavy, respectively.

\paragraph{\textbf{Visible Image Degradation Strategy}}  
Degraded visible images generation process is illustrated in \cref{fig1}. We employ a phenomenological pipeline \cite{r43} to introduce haze into source images, while rain and snow images are generated using Photoshop software utilizing masks based approach \cite{r44}. To ensure the realism of synthetic rain and snow images in the AWMM-100K dataset, we extract masks from real-world rain and snow video sequences, which are then used in the rendering process to generate degraded images. This approach preserves the natural shape, orientation, motion blur, density, and spatial structure of raindrops and snowflakes, such as linear streaks for rain and scattering or aggregation patterns for snow \cite{r44,r100}. Consequently, the synthetic data exhibit high statistical and visual realism, enabling the model to generalize effectively to real-world scenarios. As shown in \cref{fig_real_comp}, the degradations in the synthetic data of AWMM-100K closely match those observed in real degraded scenes. \cref{fig4} presents examples of multi-modality data captured in real-world scenarios.
Moreover, this benchmark provides labels for object detection. These attributes make it a versatile resource for exploring a wide range of visual tasks.

\begin{figure*}[t]
  \centering
   \includegraphics[width=1\linewidth]{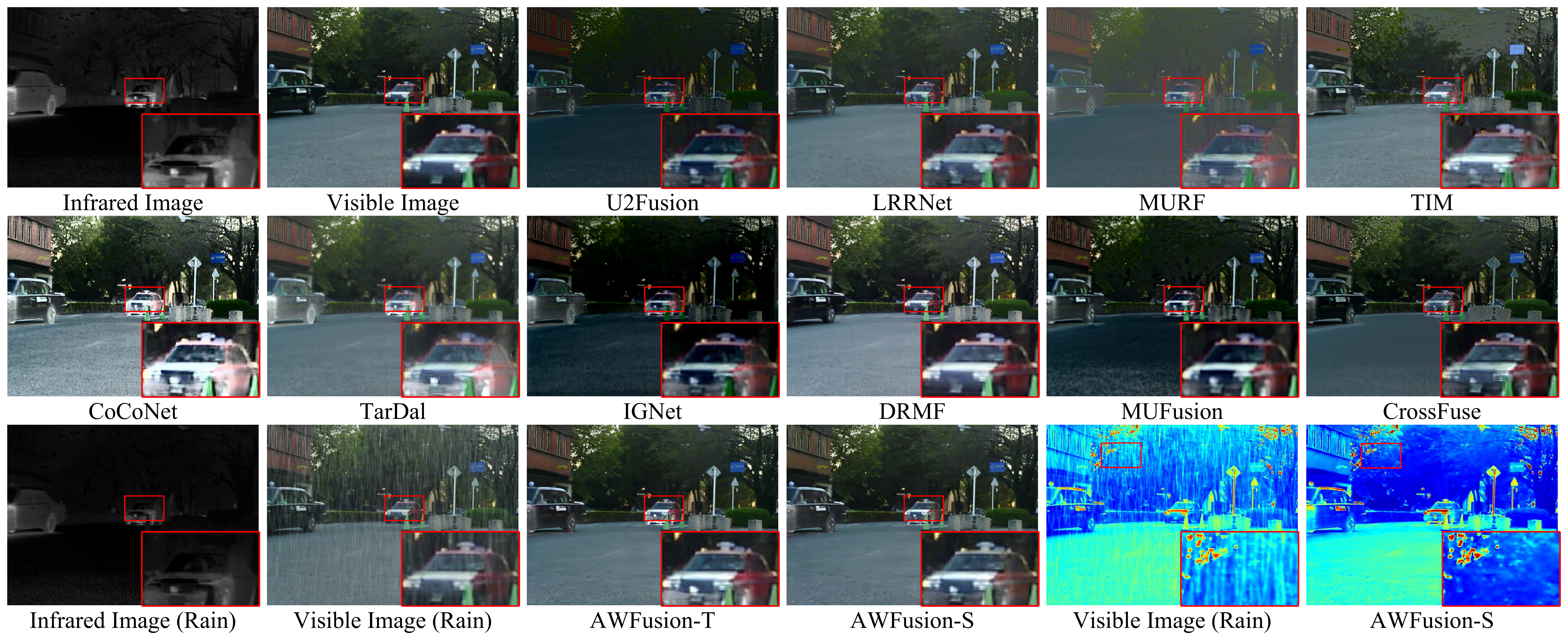}
   \caption{Comparison of fusion results obtained by the proposed algorithm under rain weather, and the results of the comparison methods under ideal condition.}
   \label{fig5}
\end{figure*}
\begin{figure*}[h!]
  \centering
   \includegraphics[width=1\linewidth]{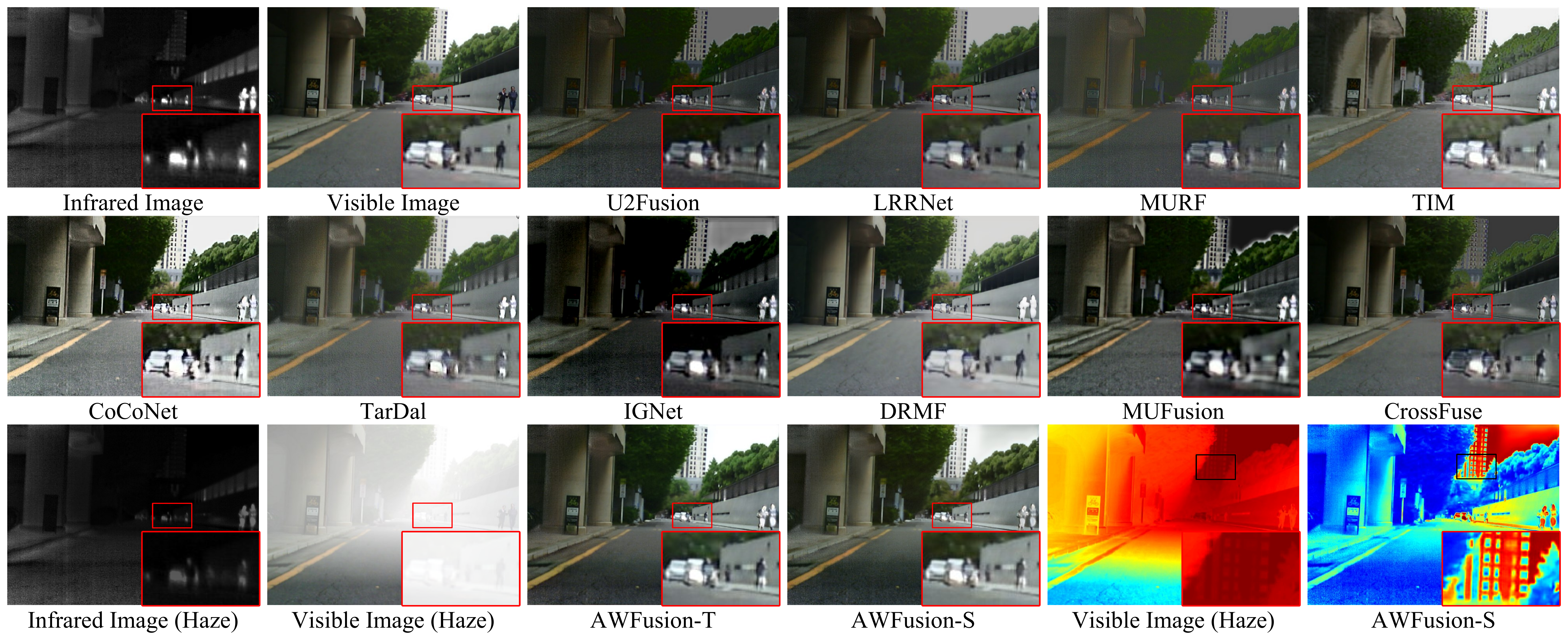}
   \caption{Comparison of fusion results obtained by the proposed algorithm under haze weather and the results of the comparison methods under ideal condition.}
   \label{fig6}
\end{figure*}

\section{Experiment}
\subsection{Experimental Setup}
We performed fusion experiments on the AWMM-100k dataset under adverse weather (rain, haze, and snow). 
The experiment included ten state-of-the-art (SOTA) IVIF algorithms (U2Fusion\cite{r2}, LRRNet\cite{r4}, MURF\cite{r36}, TIM\cite{r54}, CoCoNet\cite{r53}, TarDal\cite{r3}, IGNet\cite{r33}, DRMF\cite{r55}, MUFusion\cite{r56}, CrossFuse\cite{r57}).
We chose nine metrics to quantitatively evaluate different fusion results: Gradient-Based Fusion Performance ($Q_{G}$), Image Fusion Metric-Based on a Multiscale Scheme ($Q_{M}$), Piella’s Metric ($Q_{S}$), Chen-Blum Metric ($Q_{CB}$), Chen-Varshney Metric ($Q_{CV}$) \cite{r1,r60}, $Q^{AB/F}$ \cite{r59}, Structural Similarity Measurement ($SSIM$), the Sum of the Correlations of Differences ($SCD$)\cite{r58} and Entropy ($EN$).
For the synthesized data, the comparison method uses clean source images, while the proposed algorithm performs fusion using images affected by three different adverse weather conditions. For real scene data, both the proposed algorithm and the comparison method use the same source images. This comparison setup allows for a more effective demonstration of the performance of the proposed algorithm.

\begin{figure*}[h]
  \centering
   \includegraphics[width=1\linewidth]{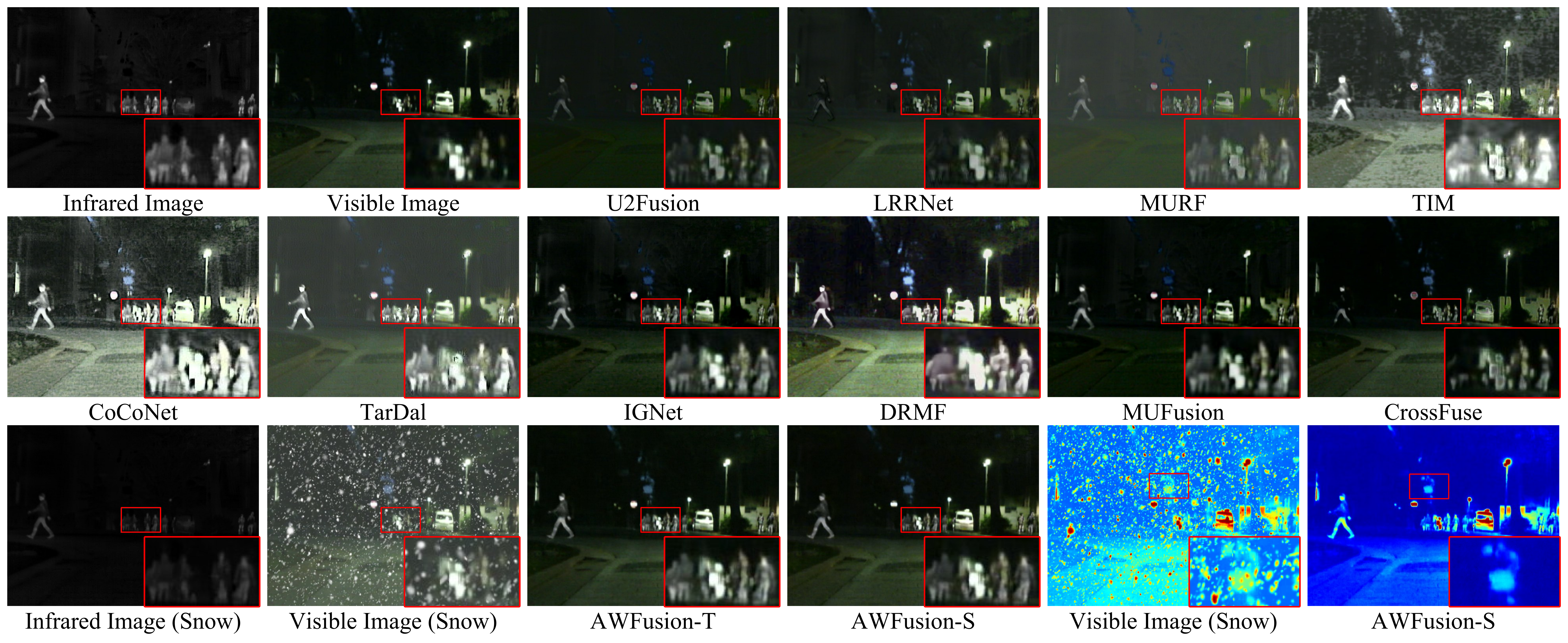}
   \caption{Comparison of fusion results obtained by the proposed algorithm under snow weather and the results of the comparison methods under ideal condition.}
   \label{fig7}
\end{figure*}
\begin{figure}[h]
  \centering
   \includegraphics[width=1\linewidth]{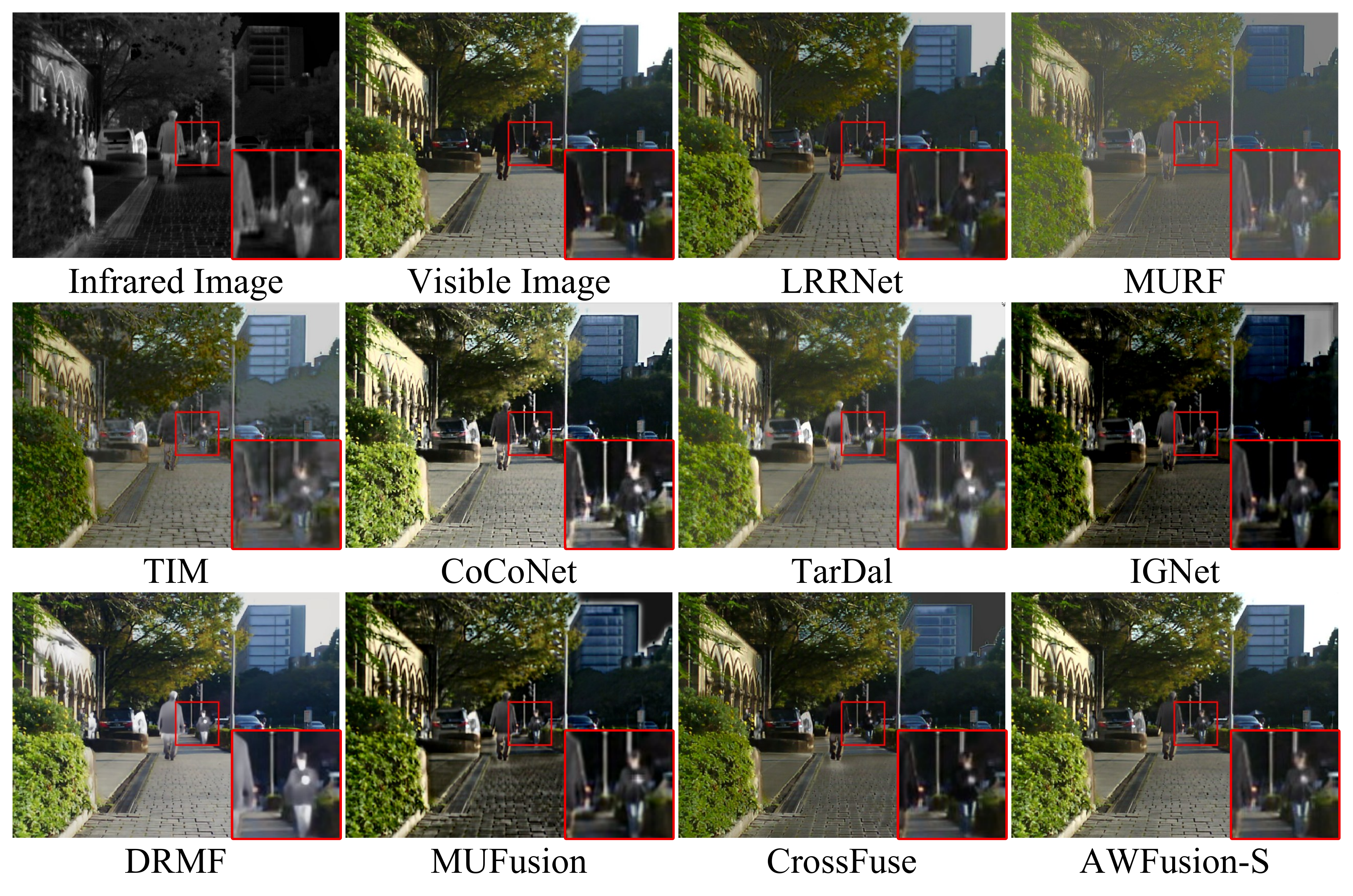}
   \caption{Comparison of the proposed algorithm and comparison methods under non-degenerate conditions.}
   \label{fig_normal}
\end{figure}
\begin{figure}[h]
  \centering
   \includegraphics[width=1\linewidth]{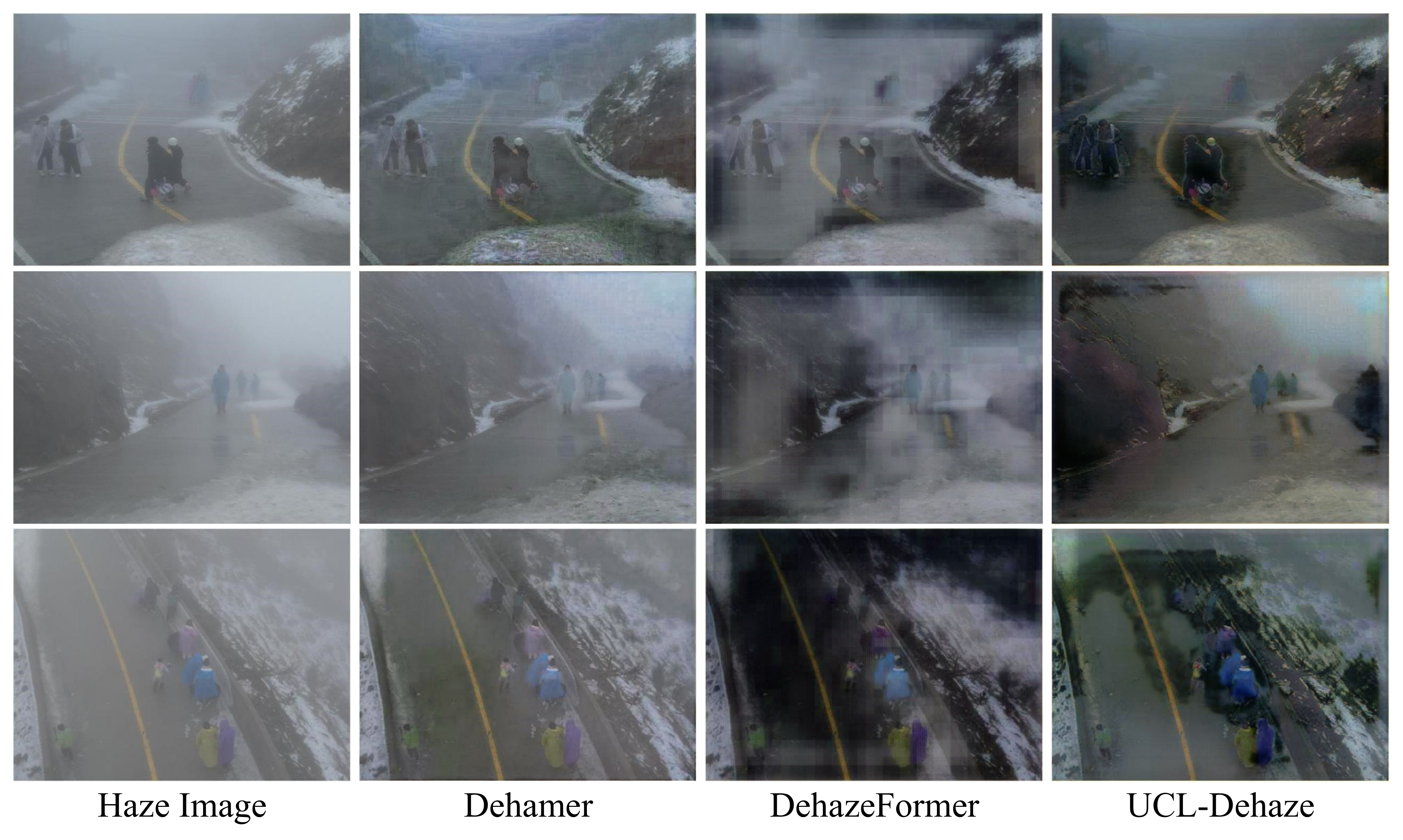}
   \caption{Results of different dehazing algorithms on real-world haze.}
   \label{fig8}
\end{figure}

\subsection{Training Details}
For the training process, three specialized teacher networks and one student network were trained under different weather conditions.
For different adverse weather conditions, we selected 1,000 pairs of multi-modality images from the AWMM-100k dataset as the training set for the teacher network. Each image was randomly cropped to a size of $160\times160$ pixels to form a batch. The initial learning rate for the Adam optimizer was set to $1 \times 10^{-3}$, and the model training was conducted over 300 iterations. During the knowledge transfer phase, the teacher network remained fixed while the student network was trained for 200 epochs on a mixed training set comprising three weather types. Furthermore, we employed two LLRR blocks, while the remaining parameters were set according to the original paper. The number of LCSC blocks iterated in the RFB, IFB, and SFB models is four, with 24 filters per block. The size of the convolutional kernel used in each convolutional layer is nine. All the experiments were conducted on a system equipped with two NVIDIA 4090 GPUs.

\begin{figure*}[t]
  \centering
   \includegraphics[width=1\linewidth]{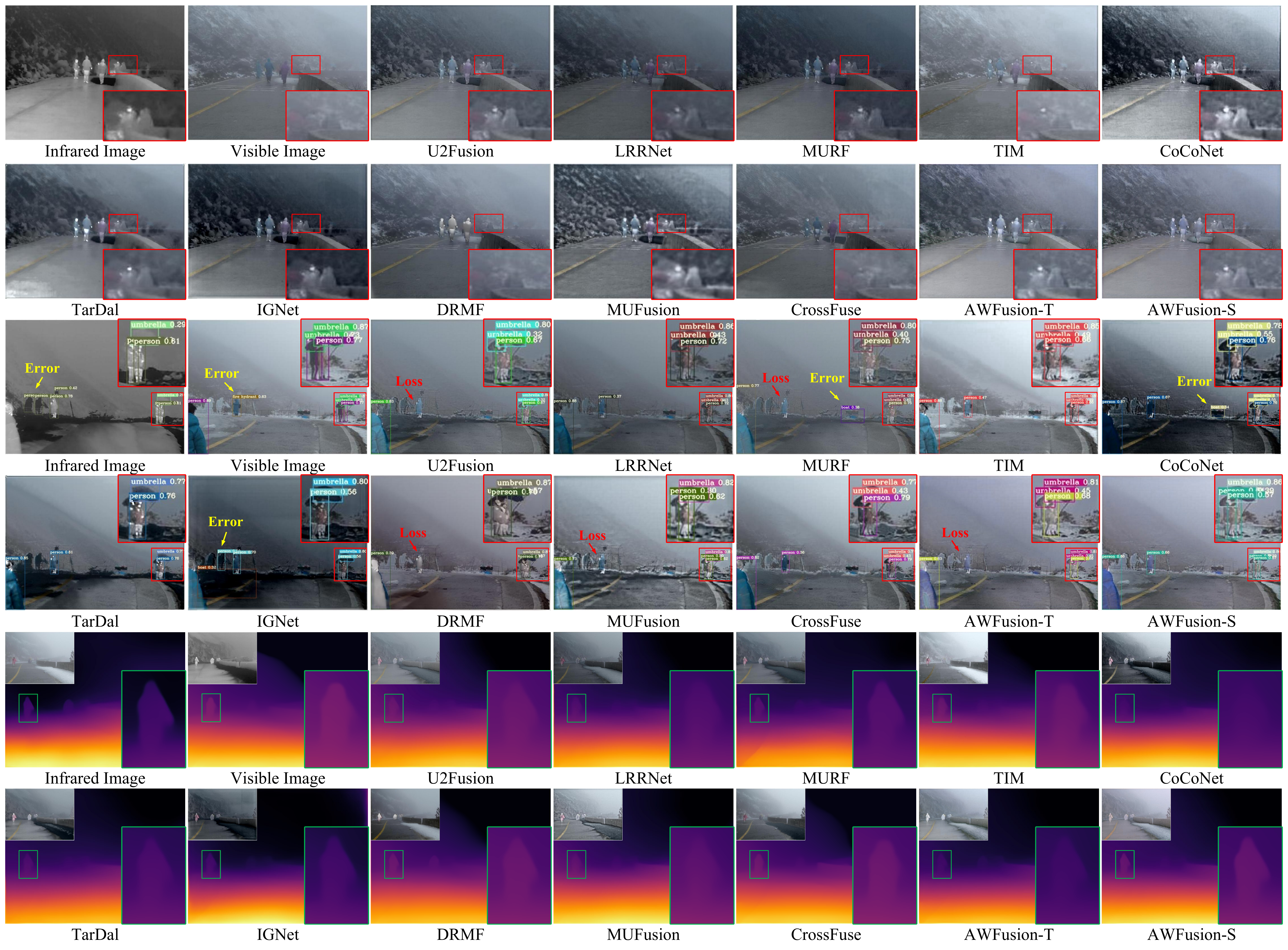}
   \caption{Comparison of fusion results of all methods in real haze scenes. The first and second rows present fusion performance comparison experiments, the third and fourth rows show object detection accuracy comparisons, and the fifth and sixth rows display depth estimation comparison experiments.}
   \label{fig9}
\end{figure*}

\begin{table*}[t]
\caption{Quantitative comparison results of all algorithms in three weather conditions (Rain, haze and snow). The comparison methods use clean source images as input, whereas the proposed method uses degraded source images. \textbf{Bold} is the best, \textcolor{red}{red} is the second and {\color[HTML]{0000FF}blue} is the third.}
\begin{adjustbox}{width=\textwidth}
\begin{tabular}{cccccccccccc}
\hline
\multicolumn{1}{c|}{\textbf{Source}}                 & \multicolumn{1}{c|}{Method}     & \multicolumn{1}{c|}{Pub.}                & $Q_{G}$↑                           & $Q_{M}$↑                            & $Q_{S}$↑                            & $Q_{CB}$↑                           & $Q_{CV}$↓               &  $Q^{AB/F}$↑             & $SSIM$↑                          & $SCD$↑                                                    & $EN$↑                            \\ \hline
\multicolumn{1}{c|}{}                                & \multicolumn{1}{c|}{U2Fusion\cite{r2}}   & \multicolumn{1}{c|}{\textit{TPAMI 2020}} & 0.2548                        & 0.4524                        & 0.7769                        & 0.4565                        & 901.4884                        & 0.2985                        & 0.3631                        & 1.0276                        & 4.4955                        \\
\multicolumn{1}{c|}{}                                & \multicolumn{1}{c|}{LRRNet\cite{r4}}     & \multicolumn{1}{c|}{\textit{TPAMI 2023}} & 0.3194                        & 0.5046                        & 0.7003                        & 0.3593                        & 682.5369                        & 0.4250                        & 0.1708                        & 0.7897                        & 5.8903                        \\
\multicolumn{1}{c|}{}                                & \multicolumn{1}{c|}{MURF\cite{r36}}       & \multicolumn{1}{c|}{\textit{TPAMI 2023}} & 0.2570                        & 0.3839                        & 0.5732                        & 0.3218                        & 1523.9926                       & 0.2285                        & 0.2699                        & 0.9151                        & 4.5988                        \\
\multicolumn{1}{c|}{}                                & \multicolumn{1}{c|}{TIM\cite{r54}}        & \multicolumn{1}{c|}{\textit{TPAMI 2024}} & 0.2057                        & 0.3498                        & 0.4630                        & 0.3844                        & 827.1437                        & 0.3035                        & 0.1585                        & 0.9651                        & 6.1997                        \\
\multicolumn{1}{c|}{}                                & \multicolumn{1}{c|}{CoCoNet\cite{r53}}    & \multicolumn{1}{c|}{\textit{IJCV 2023}}  & 0.1526                        & 0.2232                        & 0.4021                        & 0.4530                        & 1099.7753                       & 0.2129                        & 0.1756                        & 1.3002                        & \textbf{7.6143}               \\
\multicolumn{1}{c|}{}                                & \multicolumn{1}{c|}{TarDal\cite{r3}}     & \multicolumn{1}{c|}{\textit{CVPR 2022}}  & {\color[HTML]{0000FF} 0.3563} & 0.3859                        & 0.5739                        & 0.2710                        & {\color[HTML]{0000FF} 653.6011} & 0.3826                        & 0.3053                        & 1.5556                        & 6.2024                        \\
\multicolumn{1}{c|}{}                                & \multicolumn{1}{c|}{IGNet\cite{r33}}      & \multicolumn{1}{c|}{\textit{MM 2023}}    & 0.3529                        & 0.4921                        & 0.7232                        & 0.4570                        & 919.9543                        & 0.4420                        & 0.3515                        & {\color[HTML]{0000FF} 1.5791} & 6.0812                        \\
\multicolumn{1}{c|}{}                                & \multicolumn{1}{c|}{DRMF\cite{r55}}       & \multicolumn{1}{c|}{\textit{MM 2024}}    & 0.3108                        & 0.4170                        & 0.5844                        & \textbf{0.5292}               & 929.2306                        & 0.3555                        & 0.2743                        & 0.8937                        & 6.2728                        \\
\multicolumn{1}{c|}{}                                & \multicolumn{1}{c|}{MUFusion\cite{r56}}   & \multicolumn{1}{c|}{\textit{InfFus 2023}}   & 0.3438                        & 0.4696                        & {\color[HTML]{0000FF} 0.7860} & {\color[HTML]{FF0000} 0.5102} & 1039.5903                       & 0.4199                        & \textbf{0.3959}               & 1.4061                        & 5.6906                        \\
\multicolumn{1}{c|}{\multirow{-10}{*}{GT}}           & \multicolumn{1}{c|}{CrossFuse\cite{r57}}  & \multicolumn{1}{c|}{\textit{InfFus 2024}}   & \textbf{0.4057}               & {\color[HTML]{0000FF} 0.5718} & 0.7132                        & 0.4655                        & 1840.1690                       & {\color[HTML]{FF0000} 0.4648} & {\color[HTML]{FF0000} 0.3832} & 1.0081                        & 5.9859                        \\ \hline
\multicolumn{1}{c|}{}                                & \multicolumn{1}{c|}{AWFusion-T} & \multicolumn{1}{c|}{}                    & {\color[HTML]{FF0000} 0.3582} & \textbf{0.6855}               & \textbf{0.8444}               & 0.4613                        & \textbf{464.7867}               & \textbf{0.4839}               & {\color[HTML]{0000FF} 0.3693} & \textbf{1.6311}               & {\color[HTML]{FF0000} 6.3160} \\
\multicolumn{1}{c|}{\multirow{-2}{*}{\textbf{Rain}}} & \multicolumn{1}{c|}{AWFusion-S} & \multicolumn{1}{c|}{\multirow{-2}{*}{-}} & 0.3334                        & {\color[HTML]{FF0000} 0.6428} & {\color[HTML]{FF0000} 0.8335} & {\color[HTML]{0000FF} 0.4691} & {\color[HTML]{FF0000} 476.3436} & {\color[HTML]{0000FF} 0.4618} & 0.3528                        & {\color[HTML]{FF0000} 1.6010} & {\color[HTML]{0000FF} 6.2777} \\ \hline
                                                     &                                 &                                          &                               &                               &                               &                               &                                 & \multicolumn{1}{l}{}          &                               &                               & \multicolumn{1}{l}{}          \\ \hline
\multicolumn{1}{c|}{}                                & \multicolumn{1}{c|}{U2Fusion\cite{r2}}   & \multicolumn{1}{c|}{\textit{TPAMI 2020}} & 0.3654                        & 0.4275                        & 0.7236                        & \textbf{0.5051}               & 840.3653                        & 0.4086                        & 0.3911                        & 1.1512                        & 5.7205                        \\
\multicolumn{1}{c|}{}                                & \multicolumn{1}{c|}{LRRNet\cite{r4}}     & \multicolumn{1}{c|}{\textit{TPAMI 2023}} & 0.4154                        & 0.4775                        & {\color[HTML]{0000FF} 0.7646} & 0.4455                        & 463.4354                        & {\color[HTML]{0000FF} 0.5218} & 0.3046                        & 0.8130                        & 6.6553                        \\
\multicolumn{1}{c|}{}                                & \multicolumn{1}{c|}{MURF\cite{r36}}       & \multicolumn{1}{c|}{\textit{TPAMI 2023}} & 0.2589                        & 0.3324                        & 0.6395                        & 0.4055                        & 1305.4273                       & 0.2277                        & 0.3124                        & 0.9431                        & 5.3601                        \\
\multicolumn{1}{c|}{}                                & \multicolumn{1}{c|}{TIM\cite{r54}}        & \multicolumn{1}{c|}{\textit{TPAMI 2024}} & 0.3020                        & 0.3983                        & 0.6499                        & 0.4135                        & 507.1376                        & 0.4078                        & 0.2833                        & 0.8978                        & {\color[HTML]{FF0000} 7.1141} \\
\multicolumn{1}{c|}{}                                & \multicolumn{1}{c|}{CoCoNet\cite{r53}}    & \multicolumn{1}{c|}{\textit{IJCV 2023}}  & 0.2664                        & 0.2446                        & 0.6157                        & 0.4647                        & 556.1065                        & 0.3547                        & 0.2987                        & {\color[HTML]{FF0000} 1.4288} & \textbf{7.6564}               \\
\multicolumn{1}{c|}{}                                & \multicolumn{1}{c|}{TarDal\cite{r3}}     & \multicolumn{1}{c|}{\textit{CVPR 2022}}  & 0.4109                        & 0.3832                        & 0.7308                        & 0.3710                        & \textbf{337.3376}               & 0.4366                        & {\color[HTML]{0000FF} 0.4009} & {\color[HTML]{0000FF} 1.4207} & 6.8040                        \\
\multicolumn{1}{c|}{}                                & \multicolumn{1}{c|}{IGNet\cite{r33}}      & \multicolumn{1}{c|}{\textit{MM 2023}}    & 0.3271                        & 0.4117                        & 0.5225                        & 0.4234                        & 1293.6847                       & 0.4110                        & 0.2683                        & \textbf{1.4465}               & 6.2330                        \\
\multicolumn{1}{c|}{}                                & \multicolumn{1}{c|}{DRMF\cite{r55}}       & \multicolumn{1}{c|}{\textit{MM 2024}}    & \textbf{0.5203}               & \textbf{0.6138}               & 0.7530                        & 0.4463                        & 477.8087                        & 0.5026                        & \textbf{0.4182}               & 1.0414                        & 6.5732                        \\
\multicolumn{1}{c|}{}                                & \multicolumn{1}{c|}{MUFusion\cite{r56}}   & \multicolumn{1}{c|}{\textit{InfFus 2023}}   & 0.3629                        & 0.4200                        & 0.7236                        & {\color[HTML]{FF0000} 0.4908} & 2079.8455                       & 0.4451                        & 0.3723                        & 1.2420                        & 6.5892                        \\
\multicolumn{1}{c|}{\multirow{-10}{*}{GT}}           & \multicolumn{1}{c|}{CrossFuse\cite{r57}}  & \multicolumn{1}{c|}{\textit{InfFus 2024}}   & {\color[HTML]{FF0000} 0.4542} & 0.5423                        & 0.7301                        & 0.4559                        & 2474.6507                       & 0.4847                        & 0.3850                        & 0.9437                        & 6.5648                        \\ \hline
\multicolumn{1}{c|}{}                                & \multicolumn{1}{c|}{AWFusion-T} & \multicolumn{1}{c|}{}                    & {\color[HTML]{0000FF} 0.4497} & {\color[HTML]{FF0000} 0.5889} & \textbf{0.7949}               & {\color[HTML]{0000FF} 0.4838} & {\color[HTML]{FF0000} 374.9215} & \textbf{0.5585}               & {\color[HTML]{FF0000} 0.4078} & 1.2209                        & {\color[HTML]{0000FF} 6.9558} \\
\multicolumn{1}{c|}{\multirow{-2}{*}{\textbf{Haze}}} & \multicolumn{1}{c|}{AWFusion-S} & \multicolumn{1}{c|}{\multirow{-2}{*}{-}} & 0.4283                        & {\color[HTML]{0000FF} 0.5471} & {\color[HTML]{FF0000} 0.7738} & 0.4610                        & {\color[HTML]{0000FF} 462.8585} & {\color[HTML]{FF0000} 0.5322} & 0.3901                        & 1.1719                        & 6.9469                        \\ \hline
                                                     &                                 &                                          &                               &                               &                               &                               &                                 & \multicolumn{1}{l}{}          &                               &                               & \multicolumn{1}{l}{}          \\ \hline
\multicolumn{1}{c|}{}                                & \multicolumn{1}{c|}{U2Fusion\cite{r2}}   & \multicolumn{1}{c|}{\textit{TPAMI 2020}} & 0.3035                        & 0.4330                        & {\color[HTML]{0000FF} 0.7349} & 0.4547                        & 1031.8395                       & 0.3606                        & 0.3666                        & 1.0589                        & 5.1796                        \\
\multicolumn{1}{c|}{}                                & \multicolumn{1}{c|}{LRRNet\cite{r4}}     & \multicolumn{1}{c|}{\textit{TPAMI 2023}} & 0.3574                        & 0.4651                        & 0.7154                        & 0.3737                        & 661.1762                        & {\color[HTML]{0000FF} 0.4719} & 0.2361                        & 0.8301                        & 6.1878                        \\
\multicolumn{1}{c|}{}                                & \multicolumn{1}{c|}{MURF\cite{r36}}       & \multicolumn{1}{c|}{\textit{TPAMI 2023}} & 0.2440                        & 0.3547                        & 0.5581                        & 0.3482                        & 1613.7692                       & 0.2180                        & 0.2676                        & 0.8499                        & 5.0984                        \\
\multicolumn{1}{c|}{}                                & \multicolumn{1}{c|}{TIM\cite{r54}}        & \multicolumn{1}{c|}{\textit{TPAMI 2024}} & 0.2433                        & 0.3626                        & 0.5181                        & 0.3739                        & 820.5891                        & 0.3465                        & 0.2105                        & 0.8977                        & 6.5951                        \\
\multicolumn{1}{c|}{}                                & \multicolumn{1}{c|}{CoCoNet\cite{r53}}    & \multicolumn{1}{c|}{\textit{IJCV 2023}}  & 0.2420                        & 0.2580                        & 0.5238                        & 0.4512                        & 828.9278                        & 0.3277                        & 0.2593                        & 1.4450                        & \textbf{7.5726}               \\
\multicolumn{1}{c|}{}                                & \multicolumn{1}{c|}{TarDal\cite{r3}}     & \multicolumn{1}{c|}{\textit{CVPR 2022}}  & 0.3511                        & 0.3718                        & 0.6238                        & 0.2903                        & {\color[HTML]{0000FF} 553.2230} & 0.4171                        & 0.3453                        & 1.4684                        & 6.4487                        \\
\multicolumn{1}{c|}{}                                & \multicolumn{1}{c|}{IGNet\cite{r33}}      & \multicolumn{1}{c|}{\textit{MM 2023}}    & 0.3268                        & 0.4577                        & 0.5938                        & 0.4324                        & 1141.8494                       & 0.4356                        & 0.2986                        & {\color[HTML]{FF0000} 1.5142} & 6.0336                        \\
\multicolumn{1}{c|}{}                                & \multicolumn{1}{c|}{DRMF\cite{r55}}       & \multicolumn{1}{c|}{\textit{MM 2024}}    & {\color[HTML]{FF0000} 0.3948} & 0.4684                        & 0.6747                        & \textbf{0.5055}               & 729.0171                        & 0.4649                        & 0.3417                        & 1.2065                        & 6.2139                        \\
\multicolumn{1}{c|}{}                                & \multicolumn{1}{c|}{MUFusion\cite{r56}}   & \multicolumn{1}{c|}{\textit{InfFus 2023}}   & 0.3543                        & 0.4336                        & 0.7342                        & {\color[HTML]{FF0000} 0.4953} & 1699.4374                       & 0.4372                        & \textbf{0.3787}               & 1.2984                        & 6.1981                        \\
\multicolumn{1}{c|}{\multirow{-10}{*}{GT}}           & \multicolumn{1}{c|}{CrossFuse\cite{r57}}  & \multicolumn{1}{c|}{\textit{InfFus 2024}}   & \textbf{0.4090}               & {\color[HTML]{0000FF} 0.5187} & 0.6995                        & 0.4567                        & 2421.3634                       & 0.4507                        & 0.3724                        & 0.9919                        & 6.1354                        \\ \hline
\multicolumn{1}{c|}{}                                & \multicolumn{1}{c|}{AWFusion-T} & \multicolumn{1}{c|}{}                    & {\color[HTML]{0000FF} 0.3580} & \textbf{0.5591}               & \textbf{0.8179}               & {\color[HTML]{0000FF} 0.4827} & \textbf{246.8077}               & \textbf{0.4985}               & {\color[HTML]{FF0000} 0.3780} & \textbf{1.5700}               & {\color[HTML]{0000FF} 6.6173} \\
\multicolumn{1}{c|}{\multirow{-2}{*}{\textbf{Snow}}} & \multicolumn{1}{c|}{AWFusion-S} & \multicolumn{1}{c|}{\multirow{-2}{*}{-}} & 0.3552                        & {\color[HTML]{FF0000} 0.5535} & {\color[HTML]{FF0000} 0.8119} & 0.4798                        & {\color[HTML]{FF0000} 270.1595} & {\color[HTML]{FF0000} 0.4937} & {\color[HTML]{0000FF} 0.3734} & {\color[HTML]{0000FF} 1.4937} & {\color[HTML]{FF0000} 6.6531} \\ \hline
\end{tabular}
\label{tab1}
\end{adjustbox}
\end{table*}

\begin{table*}[t]
\caption{Comparison of segmentation accuracies for all methods across different categories in the MSRS dataset. BANet is used as the segmentation network. The comparison methods use clean source images as input, whereas the proposed method uses degraded source images. \textbf{Bold} is the best, \textcolor{red}{red} is the second and {\color[HTML]{0000FF}blue} is the third.}
\begin{adjustbox}{width=\textwidth}
\centering
\begin{tabular}{c|c|c|cccccccccc}
\hline
Source                             & Method                       & Pub.                & Background                   & Car                          & Person                       & Bike                         & Curve                        & Car Stop                     & Guardrail                    & Color cone                   & Bump                         & mIoU                         \\ \hline
                                   & U2Fusion\cite{r2}                     & \textit{TPAMI 2020}          & 98.39                        & 92.88                        & 73.25                        & 73.03                        & 52.82                        & 73.75                        & 86.2                         & 40.31                        & 83.2                         & 74.87                        \\
                                   & LRRNet\cite{r4}                       & \textit{TPAMI 2023}          & 98.51                        & 93.18                        & 69.62                        & 73.89                        & 65.18                        & 77.95                        & 84.59                        & 35.95                        & 88.96                        & 76.43                        \\
                                   & MURF\cite{r36}                         & \textit{TPAMI 2023}          & 99.13                        & 95.56                        & {\color[HTML]{0000FF} 83.54} & 86.65                        & 76.27                        & 88.21                        & {\color[HTML]{0000FF} 93.48} & 76.25                        & 90.59                        & 87.74                        \\
                                   & TIM\cite{r54}                          & \textit{TPAMI 2024}          & 98.22                        & 91.17                        & 67.94                        & 69.4                         & 59.53                        & 67.74                        & 83.42                        & 30.59                        & 85.59                        & 72.62                        \\
                                   & CoCoNet\cite{r53}                      & \textit{IJCV 2023}           & 98.3                         & 92.03                        & 69.31                        & 70.66                        & 57.72                        & 73.62                        & 87.49                        & 29.44                        & 84.9                         & 73.72                        \\
                                   & TarDal\cite{r3}                       & \textit{CVPR 2022}           & 98.39                        & 92.29                        & 70                           & 68.47                        & 62.07                        & 74.59                        & 87.9                         & 35.65                        & 85.42                        & 75.01                        \\
                                   & IGNet\cite{r33}                        & \textit{MM 2023}             & 99.12                        & 95.61                        & 83.09                        & 87.22                        & 74.97                        & 87.96                        & 93.1                         & 76.13                        & 90.9                         & 87.6                         \\
                                   & DRMF\cite{r55}                         & \textit{MM 2024}             & 99.16                        & 95.14                        & 82.36                        & {\color[HTML]{0000FF} 88.2}  & 84.27                        & {\color[HTML]{FF0000} 90.68} & 91.67                        & {\color[HTML]{FF0000} 80.35} & {\color[HTML]{0000FF} 92.08} & {\color[HTML]{0000FF} 89.32} \\
                                   & MUFusion\cite{r56}                     & \textit{InfFus 2023}         & {\color[HTML]{0000FF} 99.23} & 95.17                        & {\color[HTML]{FF0000} 83.84} & {\color[HTML]{FF0000} 88.43} & 84.1                         & 88.9                         & 92.94                        & \textbf{80.7}                & 86.12                        & 88.82                        \\
\multirow{-10}{*}{GT}              & CrossFuse\cite{r57}                    & \textit{InfFus 2024}         & 98.04                        & 90.3                         & 67.43                        & 68.51                        & 58.58                        & 71.81                        & 68.13                        & 29.92                        & 68.7                         & 69.04                        \\ \hline
\multicolumn{1}{l|}{{Rain}} &                              &                     & \textbf{99.32}               & \textbf{96.49}               & \textbf{84.77}               & \textbf{88.76}               & {\color[HTML]{FF0000} 86.35} & \textbf{90.45}               & \textbf{93.91}               & {\color[HTML]{0000FF} 78.45} & 91.82                        & \textbf{90.04}               \\
\multicolumn{1}{l|}{{Haze}} &                              &                     & 99.19                        & {\color[HTML]{0000FF} 95.69} & 81.51                        & 81.94                        & {\color[HTML]{0000FF} 85.21} & 89.5                         & 91.57                        & 72.73                        & \textbf{93.57}               & 87.88                        \\
\multicolumn{1}{l|}{{Snow}} & \multirow{-3}{*}{AWFusion-S} & \multirow{-3}{*}{-} & {\color[HTML]{FF0000} 99.25} & {\color[HTML]{FF0000} 95.92} & 81.24                        & 86.4                         & \textbf{86.77}               & {\color[HTML]{0000FF} 90.36} & {\color[HTML]{FF0000} 93.74} & 78.04                        & {\color[HTML]{FF0000} 92.32} & {\color[HTML]{FF0000} 89.34} \\ \hline
\end{tabular}
\end{adjustbox}
\label{tab2}
\end{table*}

\begin{table*}[t]
\caption{Comparison of detection accuracy of all methods on M3FD dataset. We used yolov7 as a detector. \textbf{Bold} is the best, \textcolor{red}{red} is the second and {\color[HTML]{0000FF}blue} is the third.}
\begin{adjustbox}{width=\textwidth}
\centering
\begin{tabular}{c|c|c|cccccccc}
\hline
Source                & Method     & Pub.                              & People                       & Car                          & Bus                          & Lamp                         & Motorcycle                   & Truck                        & {mAP@0.5} & mAP@{[}0.5:0.95{]} \\ \hline
                      & U2Fusion\cite{r2}   & \textit{TPAMI 2020}               & \textbf{0.816}               & 0.903                        & 0.901                        & { 0.716} & {\color[HTML]{0000FF} 0.711} & { 0.801} & {\color[HTML]{0000FF} 0.808}        & {\color[HTML]{FF0000} 0.512}                    \\
                      & LRRNet\cite{r4}     & \textit{TPAMI 2023}                        &  0.785 & {\color[HTML]{FF0000} 0.909} & {\color[HTML]{FF0000} 0.912} & {\color[HTML]{0000FF} 0.764} & {\color[HTML]{FF0000} 0.721} & {\color[HTML]{FF0000} 0.808} & {\color[HTML]{FF0000} 0.817}        & 0.484                                           \\
                      & MURF\cite{r36}       & \textit{TPAMI 2023} & {\color[HTML]{FF0000} 0.815} & 0.885                        & 0.881                        & 0.711                        & 0.675                        & 0.8                          & 0.794                               & 0.498                                           \\
                      & TIM\cite{r54}        & \textit{TPAMI 2024}                        & 0.731                        & 0.898                        & 0.878                        & 0.657                        & 0.656                        & 0.769                        & 0.765                               & 0.48                                            \\
                      & CoCoNet\cite{r53}    & \textit{IJCV 2023}                         & 0.791                        & 0.889                        & 0.889                        & 0.696                        & 0.641                        & 0.762                        & 0.778                               & 0.49                                            \\
                      & TarDal\cite{r3}     & \textit{CVPR 2022}                         & 0.792                        & 0.877                        & 0.857                        & 0.585                        & 0.62                         & 0.713                        & 0.741                               & 0.462                                           \\
                      & IGNet\cite{r33}      & \textit{MM 2023}                           & 0.793                        & 0.859                        & 0.836                        & 0.532                        & 0.554                        & 0.721                        & 0.716                               & 0.439                                           \\
                      & DRMF\cite{r55}       & \textit{MM 2024}                           & 0.699                        & 0.897                        & {0.896} & {\color[HTML]{FF0000} 0.787} & 0.697                        & 0.784                        & 0.793                               & 0.494                                           \\
                      & MUFusion\cite{r56}   &  \textit{InfFus 2023}   & {\color[HTML]{0000FF} 0.803} & {\color[HTML]{0000FF} 0.906} & {\color[HTML]{0000FF} 0.908} & 0.734                        & { 0.694} & {\color[HTML]{0000FF} 0.802} & {\color[HTML]{0000FF} 0.808}        & {\color[HTML]{0000FF} 0.507}                    \\
                      & CrossFuse\cite{r57}  & \textit{InfFus 2024}                          & 0.692                        & 0.891                        & 0.871                        & 0.671                        & 0.674                        & 0.771                        & 0.762                               & 0.473                                           \\ \cline{2-11} 
\multirow{-11}{*}{GT} & AWFusion-S & -                                 & 0.791               & \textbf{0.922}               & \textbf{0.93}                & \textbf{0.804}               & \textbf{0.726}               & \textbf{0.821}               & \textbf{0.832}                      & \textbf{0.534}                                  \\ \hline
\end{tabular}
\end{adjustbox}
\label{tab3}
\end{table*}

\subsection{Qualitative Comparison}
\subsubsection{Comparison on synthetic data}
\cref{fig5,fig6,fig7} present the fusion results of different algorithms on the synthetic dataset. Notably, only the proposed algorithm performs fusion under adverse weather conditions, while the comparison methods are fused in ideal scenes. Additionally, we include pseudo-color images of the disturbed visible images and the fusion results from the student model to further demonstrate the effectiveness of the proposed algorithm in mitigating severe weather interference. AWFusion-T and AWFusion-S represent the teacher and student models of the proposed algorithm, respectively. As shown in \cref{fig5,fig6,fig7}, all comparison methods achieve good fusion results in ideal scenes. Four methods, including U2Fusion, LRRNet, MURF, and IGNet, produce fusion results with low contrast. In contrast, CoCoNet and MUFusion enhance the source image features, leading to some redundancy and loss of detail in the thermal radiation information in the fusion results. In daytime scenes, the TIM, TarDal, and DRMF methods effectively retain the texture information from the visible images. However, in nighttime scenes, the TarDal method exhibits a loss of detail. Additionally, both MUFusion and CrossFuse show poor color retention of the sky, introducing excessive infrared information, which is not compatible with the human visual system.

Under severe weather interference, the challenge not only arises from weather-related artifacts on the visible images but also from the low contrast of infrared images and the loss of thermal radiation information. By examining the fusion results of the proposed algorithm in adverse weather conditions, we observe that it performs effective, high-quality fusion by removing interfering information from the visible image and enhancing weak targets in the scene. Additionally, in the red zoomed-in areas, the algorithm not only recovers a clean scene but also facilitates multi-modality information interactions, integrating salient details from different modalities. 
This results in superior performance compared to most of the comparison methods, particularly in terms of overall contrast and detail preservation. Furthermore, the pseudo-colored images show that the proposed algorithm effectively removes interference without smoothing local texture information, which is crucial for demonstrating the restoration performance of algorithm.
Since the student model exhibits stronger generalization ability compared to the teacher model, some performance is sacrificed. From the experiments, we observe that the first noticeable effect is the weakening of color information, followed by the retention of infrared information in the sky region. However, this does not affect the semantic content of the overall image, and the student model remains close to the teacher model in terms of overall contrast.

To  demonstrate the capability of proposed method in multi-modal feature interaction and extraction, we conduct a comparison under degradation-free conditions.   As shown in \cref{fig_normal}, all methods receive identical inputs.  The results indicate that our method still maintains superior fusion performance and does not fail when applied to clean scenes, despite being trained on images affected by adverse weather. 

\subsubsection{Comparison on real-world data}

\cref{fig9}, we present the fusion results of all methods in a real haze scene. Additionally, we evaluate the fusion performance of different methods using two downstream tasks: object detection and depth estimation. We used YOLOv7 \cite{r9} as a detector and MiDas \cite{r39} as a depth estimator.
Unlike synthetic data, the input images used to comparison methods and the proposed algorithm are the same, as there is no corresponding ground truth for real scene data. Furthermore, we retrained three high-performing dehaze algorithms (Dehamer\cite{r61}, DehazeFormer\cite{r62}, UCL-Dehaze\cite{r63}) using images from the AWMM-100K to optimize the visible images for input into the fusion comparison methods. However, as seen in \cref{fig8}, all dehaze methods resulted in varying degrees of color distortion, suggesting that feeding such distorted inputs into the different fusion methods could lead to unfair comparisons. Therefore, no restoration algorithm was incorporated into the fusion method to remove interfering factors from the source images.

As shown in \cref{fig9}, in the first set of fusion comparison tests, CoCoNet, TarDal, and MUFusion effectively retain the significant thermal radiation information from the infrared map. These three methods, which tend to enhance infrared data to achieve higher contrast,  can address some challenges in real haze scenes. However, this approach excessively masks details from the visible image, preventing the recovery of clear scene details. Additionally, for pedestrians submerged in haze, DRMF and CrossFuse lose target information from the infrared image, incorrectly prioritizing the blurred pedestrian details from the visible image. In contrast, the proposed algorithm strikes an effective balance between the complementary information from different modalities, maintaining good overall scene contrast while also recovering some detail in unknown real scenes. By examining the accuracies of different methods in the object detection task, it is clear that only the AWFusion-S results successfully detect all target information in the scene. While the CoCoNet method enhances target information, this also introduces detection errors, suggesting that blind enhancement is not an effective solution to the problem of degraded fusion. The proposed algorithm effectively recovers scene details and integrates complementary multi-modality information, achieving the highest detection accuracy across all objects. For the depth estimation task, the results visualize each ability of method to extract overall scene information. While all methods preserved depth in the foreground, many, such as U2Fusion, LRRNet, CoCoNet, and MUFusion, struggled with accurate depth estimation for distant pedestrians obscured by haze. 

In summary, the proposed algorithm outperforms current SOTA methods in both fusion performance and semantic information retention when addressing real-world scene challenges, and  demonstrates excellent performance in downstream tasks. 
It demonstrates excellent performance in downstream tasks, such as object detection and depth estimation, indicating its high practical application value in real-world.

\begin{figure*}[h!]
  \centering
   \includegraphics[width=1\linewidth]{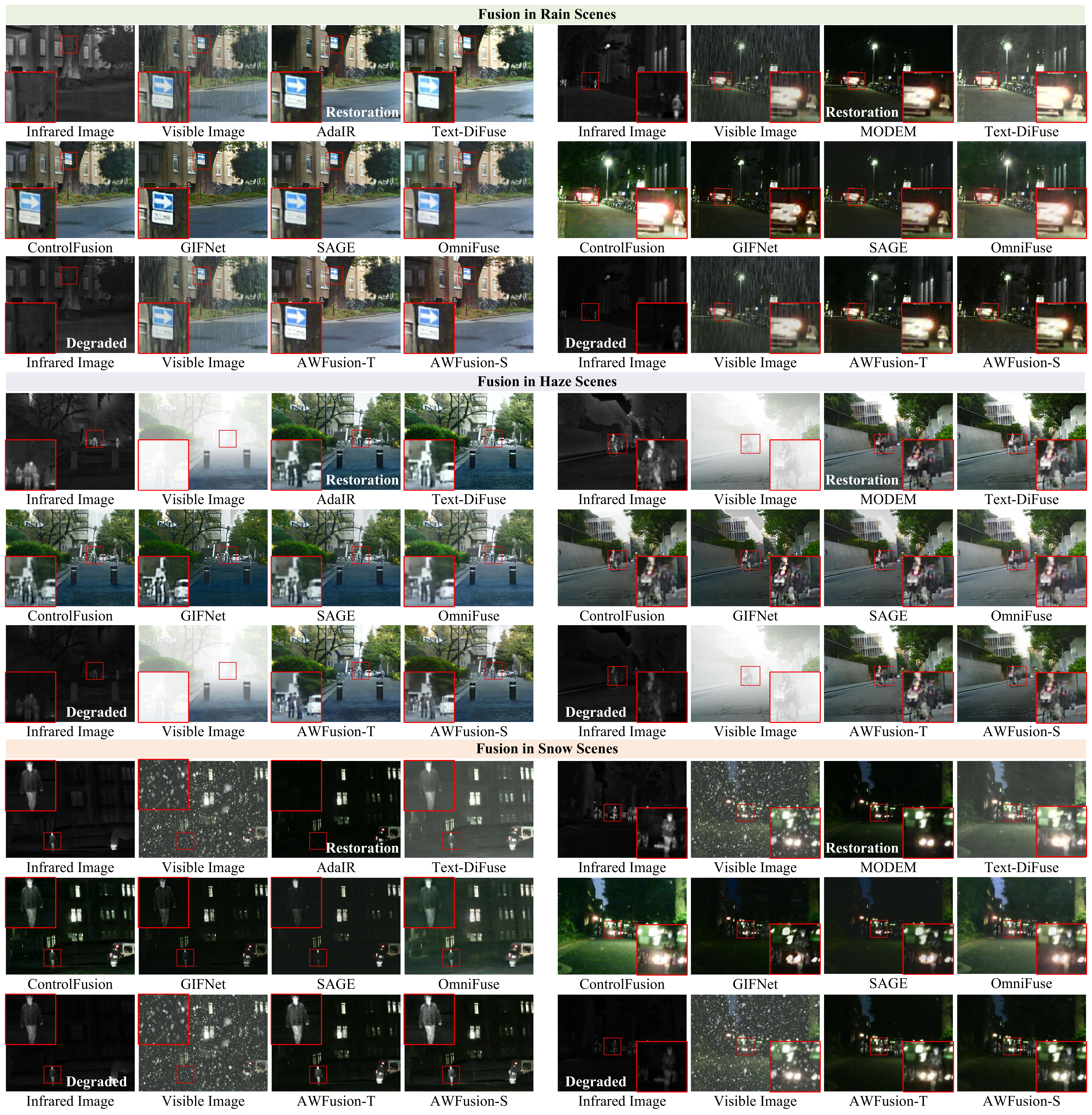}
   \caption{Comparison of different methods under three adverse weather conditions. AdaIR and MODEM are restoration methods for degraded visible images, whereas the remaining approaches are fusion methods.}
   \label{fig_restoration}
\end{figure*}
\begin{table*}[t]
\caption{Quantitative comparison results of the proposed algorithm and the combination of ``Restoration (AdaIR) + Fusion'' pipelines under three weather conditions (rain, haze and snow). \textbf{Bold} is the best, \textcolor{red}{red} is the second and {\color[HTML]{0000FF}blue} is the third. }
\begin{adjustbox}{width=\textwidth}
\begin{tabular}{cccccccccccc}
\hline
\multicolumn{1}{c|}{Methods}       & \multicolumn{1}{c|}{Pub.}                          & \multicolumn{1}{c|}{Restor.}                 &  $Q_{G}$↑                           & $Q_{M}$↑                            & $Q_{S}$↑                            & $Q_{CB}$↑                           & $Q_{CV}$↓               &  $Q^{AB/F}$↑             & $SSIM$↑                          & $SCD$↑                                                    & $EN$↑                             \\ \hline
\multicolumn{12}{c}{Rain}                                                                                                                                                                                                                                                                                                                                                                                                                \\ \hline
\multicolumn{1}{c|}{Text-DiFuse \cite{r96}}   & \multicolumn{1}{c|}{\textit{NIPS 24}}              & \multicolumn{1}{c|}{}                        & 0.2442                        & 0.3645                        & 0.5283                        & 0.3909                        & 777.6507                        & 0.1962                        & 1.2938                        & 0.3329                        & {\color[HTML]{FF0000} 6.9609} \\
\multicolumn{1}{c|}{ControlFusion \cite{r101}} & \multicolumn{1}{c|}{\textit{NIPS 25}}              & \multicolumn{1}{c|}{}                        & {\color[HTML]{FF0000} 0.3507} & 0.4750                        & 0.6901                        & 0.4361                        & 739.1370                        & 0.2996                        & 1.1971                        & {\color[HTML]{0000FF} 0.4576} & \textbf{7.0261}               \\
\multicolumn{1}{c|}{GIFNet \cite{r86}}        & \multicolumn{1}{c|}{\textit{CVPR 25}}              & \multicolumn{1}{c|}{}                        & 0.2975                        & 0.3686                        & {\color[HTML]{0000FF} 0.8040} & \textbf{0.4834}               & 672.4981                        & {\color[HTML]{0000FF} 0.3474} & {\color[HTML]{0000FF} 1.3440} & 0.3544                        & 5.5197                        \\
\multicolumn{1}{c|}{SAGE \cite{r91}}          & \multicolumn{1}{c|}{\textit{CVPR 25}}              & \multicolumn{1}{c|}{}                        & 0.2286                        & {\color[HTML]{0000FF} 0.5086} & 0.7878                        & 0.3964                        & {\color[HTML]{FF0000} 465.1593} & 0.2734                        & 1.2986                        & 0.3670                        & 5.4524                        \\
\multicolumn{1}{c|}{OmniFuse \cite{r98}}      & \multicolumn{1}{c|}{\textit{TPAMI 25}}             & \multicolumn{1}{c|}{\multirow{-5}{*}{AdaIR \cite{r102}}} & 0.2191                        & 0.4027                        & 0.5863                        & 0.3774                        & 549.3715                        & 0.1771                        & 1.0588                        & 0.3340                        & {\color[HTML]{0000FF} 6.9013} \\ \hline
\multicolumn{1}{c|}{AWFusion-T}    & \multicolumn{1}{c|}{}                              & \multicolumn{1}{c|}{}                        & \textbf{0.3580}               & \textbf{0.6855}               & \textbf{0.8444}               & {\color[HTML]{0000FF} 0.4613} & \textbf{464.7867}               & \textbf{0.3693}               & \textbf{1.6311}               & \textbf{0.4839}               & 6.3160                        \\
\multicolumn{1}{c|}{AWFusion-S}    & \multicolumn{1}{c|}{\multirow{-2}{*}{\textit{\_}}} & \multicolumn{1}{c|}{\multirow{-2}{*}{\_}}    & {\color[HTML]{0000FF} 0.3334} & {\color[HTML]{FF0000} 0.6428} & {\color[HTML]{FF0000} 0.8335} & {\color[HTML]{FF0000} 0.4691} & {\color[HTML]{0000FF} 476.3436} & {\color[HTML]{FF0000} 0.3528} & {\color[HTML]{FF0000} 1.6010} & {\color[HTML]{FF0000} 0.4618} & 6.2777                        \\ \hline
\multicolumn{12}{c}{Haze}                                                                                                                                                                                                                                                                                                                                                                                                                \\ \hline
\multicolumn{1}{c|}{Text-DiFuse \cite{r96}}   & \multicolumn{1}{c|}{\textit{NIPS 24}}              & \multicolumn{1}{c|}{}                        & 0.3137                        & 0.3463                        & 0.6622                        & 0.4391                        & 597.2099                        & 0.3035                        & \textbf{1.2850}               & 0.4021                        & \textbf{7.3050}               \\
\multicolumn{1}{c|}{ControlFusion \cite{r101}} & \multicolumn{1}{c|}{\textit{NIPS 25}}              & \multicolumn{1}{c|}{}                        & {\color[HTML]{0000FF} 0.3841} & {\color[HTML]{0000FF} 0.4032} & 0.7309                        & {\color[HTML]{0000FF} 0.4652} & 547.6077                        & 0.3331                        & 1.1490                        & {\color[HTML]{0000FF} 0.4767} & 6.9131                        \\
\multicolumn{1}{c|}{GIFNet \cite{r86}}        & \multicolumn{1}{c|}{\textit{CVPR 25}}              & \multicolumn{1}{c|}{}                        & 0.3124                        & 0.3181                        & 0.6878                        & {\color[HTML]{FF0000} 0.4784} & 847.9477                        & {\color[HTML]{0000FF} 0.3463} & {\color[HTML]{0000FF} 1.2118} & 0.3478                        & 6.2780                        \\
\multicolumn{1}{c|}{SAGE \cite{r91}}          & \multicolumn{1}{c|}{\textit{CVPR 25}}              & \multicolumn{1}{c|}{}                        & 0.3416                        & 0.3948                        & {\color[HTML]{0000FF} 0.7347} & 0.4553                        & 557.8396                        & 0.3316                        & 1.1922                        & 0.4336                        & 6.6047                        \\
\multicolumn{1}{c|}{OmniFuse \cite{r98}}      & \multicolumn{1}{c|}{\textit{TPAMI 25}}             & \multicolumn{1}{c|}{\multirow{-5}{*}{AdaIR \cite{r102}}} & 0.2638                        & 0.3448                        & 0.6483                        & 0.4226                        & {\color[HTML]{0000FF} 514.9563} & 0.2558                        & 1.0287                        & 0.3446                        & {\color[HTML]{FF0000} 7.0612} \\ \hline
\multicolumn{1}{c|}{AWFusion-T}    & \multicolumn{1}{c|}{}                              & \multicolumn{1}{c|}{}                        & \textbf{0.4497}               & \textbf{0.5889}               & \textbf{0.7949}               & \textbf{0.4838}               & \textbf{374.9215}               & \textbf{0.4078}               & {\color[HTML]{FF0000} 1.2209} & \textbf{0.5585}               & {\color[HTML]{0000FF} 6.9558} \\
\multicolumn{1}{c|}{AWFusion-S}    & \multicolumn{1}{c|}{\multirow{-2}{*}{\textit{\_}}} & \multicolumn{1}{c|}{\multirow{-2}{*}{\_}}    & {\color[HTML]{FF0000} 0.4283} & {\color[HTML]{FF0000} 0.5471} & {\color[HTML]{FF0000} 0.7738} & 0.4610                        & {\color[HTML]{FF0000} 462.8585} & {\color[HTML]{FF0000} 0.3901} & 1.1719                        & {\color[HTML]{FF0000} 0.5322} & 6.9469                        \\ \hline
\multicolumn{12}{c}{Snow}                                                                                                                                                                                                                                                                                                                                                                                                                \\ \hline
\multicolumn{1}{c|}{Text-DiFuse \cite{r96}}   & \multicolumn{1}{c|}{\textit{NIPS 24}}              & \multicolumn{1}{c|}{}                        & 0.3047                        & 0.3633                        & 0.6263                        & 0.4264                        & 502.4444                        & 0.2821                        & {\color[HTML]{FF0000} 1.5002} & 0.4130                        & \textbf{7.0752}               \\
\multicolumn{1}{c|}{ControlFusion \cite{r101}} & \multicolumn{1}{c|}{\textit{NIPS 25}}              & \multicolumn{1}{c|}{}                        & \textbf{0.4184}               & 0.4610                        & 0.7487                        & \textbf{0.4855}               & 427.3205                        & 0.3554                        & 1.3418                        & \textbf{0.5194}               & {\color[HTML]{0000FF} 7.0112} \\
\multicolumn{1}{c|}{GIFNet \cite{r86}}        & \multicolumn{1}{c|}{\textit{CVPR 25}}              & \multicolumn{1}{c|}{}                        & 0.3282                        & 0.3305                        & 0.7706                        & {\color[HTML]{FF0000} 0.4854} & 482.2309                        & {\color[HTML]{0000FF} 0.3722} & 1.3885                        & 0.3882                        & 6.0665                        \\
\multicolumn{1}{c|}{SAGE \cite{r91}}          & \multicolumn{1}{c|}{\textit{CVPR 25}}              & \multicolumn{1}{c|}{}                        & 0.3149                        & {\color[HTML]{0000FF} 0.4752} & {\color[HTML]{0000FF} 0.7910} & 0.4421                        & {\color[HTML]{0000FF} 275.8248} & 0.3474                        & 1.4294                        & 0.4585                        & 6.0712                        \\
\multicolumn{1}{c|}{OmniFuse \cite{r98}}      & \multicolumn{1}{c|}{\textit{TPAMI 25}}             & \multicolumn{1}{c|}{\multirow{-5}{*}{AdaIR \cite{r102}}} & 0.2729                        & 0.3845                        & 0.6254                        & 0.4114                        & 318.5334                        & 0.2515                        & 1.1877                        & 0.3977                        & {\color[HTML]{FF0000} 7.0768} \\ \hline
\multicolumn{1}{c|}{AWFusion-T}    & \multicolumn{1}{c|}{}                              & \multicolumn{1}{c|}{}                        & {\color[HTML]{FF0000} 0.3580} & \textbf{0.5591}               & \textbf{0.8179}               & {\color[HTML]{0000FF} 0.4827} & \textbf{246.8077}               & \textbf{0.3780}               & \textbf{1.5700}               & {\color[HTML]{FF0000} 0.4985} & 6.6173                        \\
\multicolumn{1}{c|}{AWFusion-S}    & \multicolumn{1}{c|}{\multirow{-2}{*}{\textit{\_}}} & \multicolumn{1}{c|}{\multirow{-2}{*}{\_}}    & {\color[HTML]{0000FF} 0.3552} & {\color[HTML]{FF0000} 0.5535} & {\color[HTML]{FF0000} 0.8119} & 0.4798                        & {\color[HTML]{FF0000} 270.1595} & {\color[HTML]{FF0000} 0.3734} & {\color[HTML]{0000FF} 1.4937} & {\color[HTML]{0000FF} 0.4937} & 6.6531                        \\ \hline
\end{tabular}
\label{lab_AdaIR}
\end{adjustbox}
\end{table*}

\begin{table*}[t]
\caption{Quantitative comparison results of the proposed algorithm and the combination of ``Restoration (MODEM) + Fusion'' pipelines under three weather conditions (rain, haze and snow). \textbf{Bold} is the best, \textcolor{red}{red} is the second and {\color[HTML]{0000FF}blue} is the third. }
\begin{adjustbox}{width=\textwidth}
\begin{tabular}{cccccccccccc}
\hline
\multicolumn{1}{c|}{Methods}       & \multicolumn{1}{c|}{Pub.}                          & \multicolumn{1}{c|}{Restor.}                 & $Q_{G}$↑                           & $Q_{M}$↑                            & $Q_{S}$↑                            & $Q_{CB}$↑                           & $Q_{CV}$↓               &  $Q^{AB/F}$↑             & $SSIM$↑                          & $SCD$↑                                                    & $EN$↑                            \\ \hline
\multicolumn{12}{c}{Rain}                                                                                                                                                                                                                                                                                                                                                                                                                \\ \hline
\multicolumn{1}{c|}{Text-DiFuse \cite{r96}}   & \multicolumn{1}{c|}{\textit{NIPS 24}}              & \multicolumn{1}{c|}{}                        & 0.2485                        & 0.3622                        & 0.5280                        & 0.4076                        & 770.4863                        & 0.2000                        & 1.2776                        & 0.3384                        & {\color[HTML]{FF0000} 6.9881} \\
\multicolumn{1}{c|}{ControlFusion \cite{r101}} & \multicolumn{1}{c|}{\textit{NIPS 25}}              & \multicolumn{1}{c|}{}                        & \textbf{0.3809}               & 0.4677                        & 0.6769                        & 0.4420                        & 740.8956                        & 0.2956                        & 1.1693                        & {\color[HTML]{0000FF} 0.4531} & \textbf{7.0698}               \\
\multicolumn{1}{c|}{GIFNet \cite{r86}}        & \multicolumn{1}{c|}{\textit{CVPR 25}}              & \multicolumn{1}{c|}{}                        & 0.3075                        & 0.3747                        & {\color[HTML]{0000FF} 0.8062} & \textbf{0.4804}               & 670.4288                        & {\color[HTML]{0000FF} 0.3512} & {\color[HTML]{0000FF} 1.3463} & 0.3650                        & 5.5507                        \\
\multicolumn{1}{c|}{SAGE \cite{r91}}          & \multicolumn{1}{c|}{\textit{CVPR 25}}              & \multicolumn{1}{c|}{}                        & 0.2355                        & {\color[HTML]{0000FF} 0.5356} & 0.7925                        & 0.4072                        & {\color[HTML]{FF0000} 465.1666} & 0.2872                        & 1.3023                        & 0.3793                        & 5.5068                        \\
\multicolumn{1}{c|}{OmniFuse \cite{r98}}      & \multicolumn{1}{c|}{\textit{TPAMI 25}}             & \multicolumn{1}{c|}{\multirow{-5}{*}{MODEM \cite{r103}}} & 0.2248                        & 0.4068                        & 0.5903                        & 0.3912                        & 545.6923                        & 0.1845                        & 1.0391                        & 0.3434                        & {\color[HTML]{0000FF} 6.9319} \\ \hline
\multicolumn{1}{c|}{AWFusion-T}    & \multicolumn{1}{c|}{}                              & \multicolumn{1}{c|}{}                        & {\color[HTML]{FF0000} 0.3580} & \textbf{0.6855}               & \textbf{0.8444}               & {\color[HTML]{0000FF} 0.4613} & \textbf{464.7867}               & \textbf{0.3693}               & \textbf{1.6311}               & \textbf{0.4839}               & 6.3160                        \\
\multicolumn{1}{c|}{AWFusion-S}    & \multicolumn{1}{c|}{\multirow{-2}{*}{\textit{\_}}} & \multicolumn{1}{c|}{\multirow{-2}{*}{\_}}    & {\color[HTML]{0000FF} 0.3334} & {\color[HTML]{FF0000} 0.6428} & {\color[HTML]{FF0000} 0.8335} & {\color[HTML]{FF0000} 0.4691} & {\color[HTML]{0000FF} 476.3436} & {\color[HTML]{FF0000} 0.3528} & {\color[HTML]{FF0000} 1.6010} & {\color[HTML]{FF0000} 0.4618} & 6.2777                        \\ \hline
\multicolumn{12}{c}{Haze}                                                                                                                                                                                                                                                                                                                                                                                                                \\ \hline
\multicolumn{1}{c|}{Text-DiFuse \cite{r96}}   & \multicolumn{1}{c|}{\textit{NIPS 24}}              & \multicolumn{1}{c|}{}                        & 0.3513                        & 0.3752                        & 0.7024                        & 0.4728                        & {\color[HTML]{0000FF} 516.7434} & 0.3380                        & \textbf{1.3969}               & 0.4577                        & \textbf{7.3805}               \\
\multicolumn{1}{c|}{ControlFusion \cite{r101}} & \multicolumn{1}{c|}{\textit{NIPS 25}}              & \multicolumn{1}{c|}{}                        & {\color[HTML]{FF0000} 0.4402} & 0.4761                        & 0.7560                        & {\color[HTML]{FF0000} 0.4981} & 553.4258                        & 0.3673                        & 1.2045                        & {\color[HTML]{FF0000} 0.5511} & {\color[HTML]{FF0000} 7.1785} \\
\multicolumn{1}{c|}{GIFNet \cite{r86}}        & \multicolumn{1}{c|}{\textit{CVPR 25}}              & \multicolumn{1}{c|}{}                        & 0.3530                        & 0.3366                        & 0.7344                        & \textbf{0.5007}               & 1090.9966                       & 0.3834                        & {\color[HTML]{0000FF} 1.2959} & 0.4085                        & 6.3176                        \\
\multicolumn{1}{c|}{SAGE \cite{r91}}          & \multicolumn{1}{c|}{\textit{CVPR 25}}              & \multicolumn{1}{c|}{}                        & 0.4026                        & {\color[HTML]{0000FF} 0.5124} & {\color[HTML]{FF0000} 0.7906} & 0.4826                        & 626.5782                        & {\color[HTML]{FF0000} 0.3935} & {\color[HTML]{FF0000} 1.3033} & 0.5216                        & 6.6250                        \\
\multicolumn{1}{c|}{OmniFuse \cite{r98}}      & \multicolumn{1}{c|}{\textit{TPAMI 25}}             & \multicolumn{1}{c|}{\multirow{-5}{*}{MODEM \cite{r103}}} & 0.2866                        & 0.3705                        & 0.6982                        & 0.4621                        & 529.6896                        & 0.2834                        & 1.0791                        & 0.4028                        & {\color[HTML]{0000FF} 7.1742} \\ \hline
\multicolumn{1}{c|}{AWFusion-T}    & \multicolumn{1}{c|}{}                              & \multicolumn{1}{c|}{}                        & \textbf{0.4497}               & \textbf{0.5889}               & \textbf{0.7949}               & {\color[HTML]{0000FF} 0.4838} & \textbf{374.9215}               & \textbf{0.4078}               & 1.2209                        & \textbf{0.5585}               & 6.9558                        \\
\multicolumn{1}{c|}{AWFusion-S}    & \multicolumn{1}{c|}{\multirow{-2}{*}{\textit{\_}}} & \multicolumn{1}{c|}{\multirow{-2}{*}{\_}}    & {\color[HTML]{0000FF} 0.4283} & {\color[HTML]{FF0000} 0.5471} & {\color[HTML]{0000FF} 0.7738} & 0.4610                        & {\color[HTML]{FF0000} 462.8585} & {\color[HTML]{0000FF} 0.3901} & 1.1719                        & {\color[HTML]{0000FF} 0.5322} & 6.9469                        \\ \hline
\multicolumn{12}{c}{Snow}                                                                                                                                                                                                                                                                                                                                                                                                                \\ \hline
\multicolumn{1}{c|}{Text-DiFuse \cite{r96}}   & \multicolumn{1}{c|}{\textit{NIPS 24}}              & \multicolumn{1}{c|}{}                        & 0.3089                        & 0.3672                        & 0.6289                        & 0.4312                        & 492.1872                        & 0.2848                        & {\color[HTML]{0000FF} 1.4871} & 0.4207                        & \textbf{7.0754}               \\
\multicolumn{1}{c|}{ControlFusion \cite{r101}} & \multicolumn{1}{c|}{\textit{NIPS 25}}              & \multicolumn{1}{c|}{}                        & \textbf{0.4150}               & 0.4619                        & 0.7486                        & {\color[HTML]{0000FF} 0.4816} & 422.5371                        & 0.3538                        & 1.3377                        & \textbf{0.5220}               & {\color[HTML]{0000FF} 7.0211} \\
\multicolumn{1}{c|}{GIFNet \cite{r86}}        & \multicolumn{1}{c|}{\textit{CVPR 25}}              & \multicolumn{1}{c|}{}                        & 0.3227                        & 0.3350                        & 0.7693                        & {\color[HTML]{FF0000} 0.4821} & 487.3009                        & {\color[HTML]{0000FF} 0.3665} & 1.3761                        & 0.3874                        & 6.0588                        \\
\multicolumn{1}{c|}{SAGE \cite{r91}}          & \multicolumn{1}{c|}{\textit{CVPR 25}}              & \multicolumn{1}{c|}{}                        & 0.3172                        & {\color[HTML]{0000FF} 0.4913} & {\color[HTML]{0000FF} 0.7937} & 0.4453                        & {\color[HTML]{FF0000} 262.5657} & 0.3486                        & 1.4287                        & 0.4653                        & 6.0729                        \\
\multicolumn{1}{c|}{OmniFuse \cite{r98}}      & \multicolumn{1}{c|}{\textit{TPAMI 25}}             & \multicolumn{1}{c|}{\multirow{-5}{*}{MODEM \cite{r103}}} & 0.2763                        & 0.3902                        & 0.6289                        & 0.4182                        & 311.3442                        & 0.2540                        & 1.1914                        & 0.4052                        & {\color[HTML]{FF0000} 7.0759} \\ \hline
\multicolumn{1}{c|}{AWFusion-T}    & \multicolumn{1}{c|}{}                              & \multicolumn{1}{c|}{}                        & {\color[HTML]{FF0000} 0.3580} & \textbf{0.5591}               & \textbf{0.8179}               & \textbf{0.4827}               & \textbf{246.8077}               & \textbf{0.3780}               & \textbf{1.5700}               & {\color[HTML]{FF0000} 0.4985} & 6.6173                        \\
\multicolumn{1}{c|}{AWFusion-S}    & \multicolumn{1}{c|}{\multirow{-2}{*}{\textit{\_}}} & \multicolumn{1}{c|}{\multirow{-2}{*}{\_}}    & {\color[HTML]{0000FF} 0.3552} & {\color[HTML]{FF0000} 0.5535} & {\color[HTML]{FF0000} 0.8119} & 0.4798                        & {\color[HTML]{0000FF} 270.1595} & {\color[HTML]{FF0000} 0.3734} & {\color[HTML]{FF0000} 1.4937} & {\color[HTML]{0000FF} 0.4937} & 6.6531                        \\ \hline
\end{tabular}
\label{label_MODEM}
\end{adjustbox}
\end{table*}

\subsection{Quantitative Comparison}
\subsubsection{Fusion Performance Comparison}
We selected 50 pairs of infrared and visible source images from the AWMM 100k dataset for each of the three types of adverse weather: rain, haze, and snow, as the test set. The ground truth corresponding to the degraded images was used as input for all comparison methods. 
\cref{tab1} presents the results of all methods across nine objective metrics, with the first column indicating the source image types used by each method. The top three scores for each metric are highlighted. Notably, \cref{tab1} shows that the proposed method, even when performing fusion under adverse weather conditions, outperforms comparison methods operating under ideal conditions.
In rain, haze, and snow, the proposed teacher model ranked in the top three across more than eight evaluation metrics.  Meanwhile, the student model achieved top-three rankings on seven, four, and seven metrics under the respective conditions.

By specifically analyzing the scores of each metric, we can draw the following conclusions:

\noindent(1) In the image feature-based metrics $Q_{G}$ and $Q_{M}$, which are primarily computed using the Sobel edge operator and two-level Haar wavelet, respectively, these metrics assess the edge information in the fusion results. The proposed algorithm achieves high scores in both, indicating that AWFusion preserves the edge structure while effectively removing degradation (such as haze) that significantly obscures edges.

\noindent(2) Analyzing the $SSIM$ and $SCD$ metrics, it is worth noting that AWFusion-S performs slightly worse in rain and haze scenes compared to snow on the $SSIM$ metric. This discrepancy arises because rain and haze are continuously distributed degradations, while snow is more localized. As a result, snow is slightly easier to process than rain and haze. 
The proposed algorithm is not among the top three in the $SCD$ metric for haze scenes. This metric evaluates image quality based on texture and local structural changes, suggesting that the performance of algorithm  in extracting local texture under haze is somewhat weaker than in other conditions.

\noindent(3) For the metrics $Q_{CB}$ and $Q_{CV}$, which are inspired by human perception, the proposed method performs well. This aligns with the results shown in \cref{fig5,fig6,fig7} and \cref{fig9}, where the method effectively integrates multi-modality information while eliminating interference.


\subsubsection{Semantic Segmentation Comparison}

The proposed all-weather MMIF algorithms are designed to support real-world applications, such as autonomous driving, surveillance, and UAV navigation. In these scenarios, the goal is not just to generate visually enhanced images but to enable effective decision-making and analysis (e.g., object detection, semantic segmentation) from the fused images. To comprehensively evaluate the fusion performance under three types of severe weather, we introduced a downstream task.
In \cref{tab2}, we compare the performance of various algorithms using BANet \cite{r65} as the segmentation network. Unlike comparison methods, which receive clean source images as input, the proposed algorithm performs fusion under severe weather conditions. As shown in \cref{tab2}, the proposed algorithm ranks first in mean intersection-over-union (mIoU) for rain scenes and second for snow scenes. 
Although the mIoU score for the haze scenario is not in the top three, it still secures fifth place. These experiments demonstrate that the proposed algorithm retains high-level semantic information in scenes with interference and accurately recognizes key targets such as roads, pedestrians, and vehicles.

\begin{figure*}[t]
  \centering
   \includegraphics[width=1\linewidth]{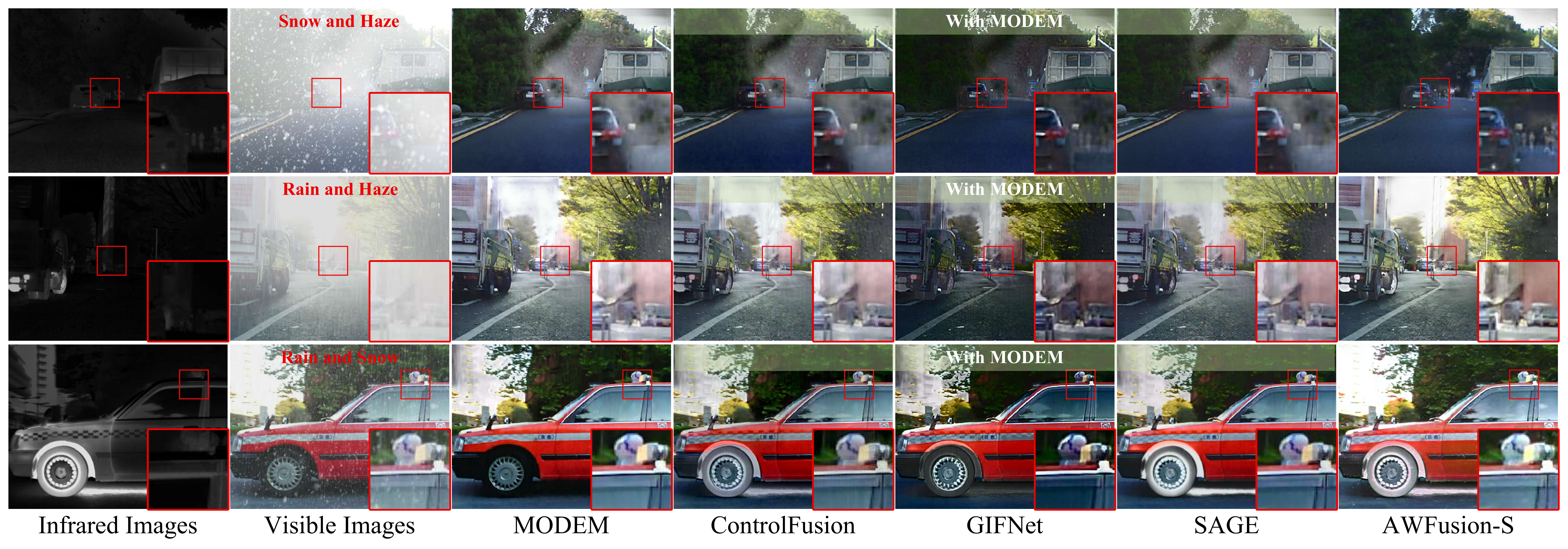}
   \caption{Comparison of fusion results of different methods under multiple adverse weather disturbances.}
   \label{fig11}
\end{figure*}
\begin{figure}[h]
  \centering
   \includegraphics[width=1\linewidth]{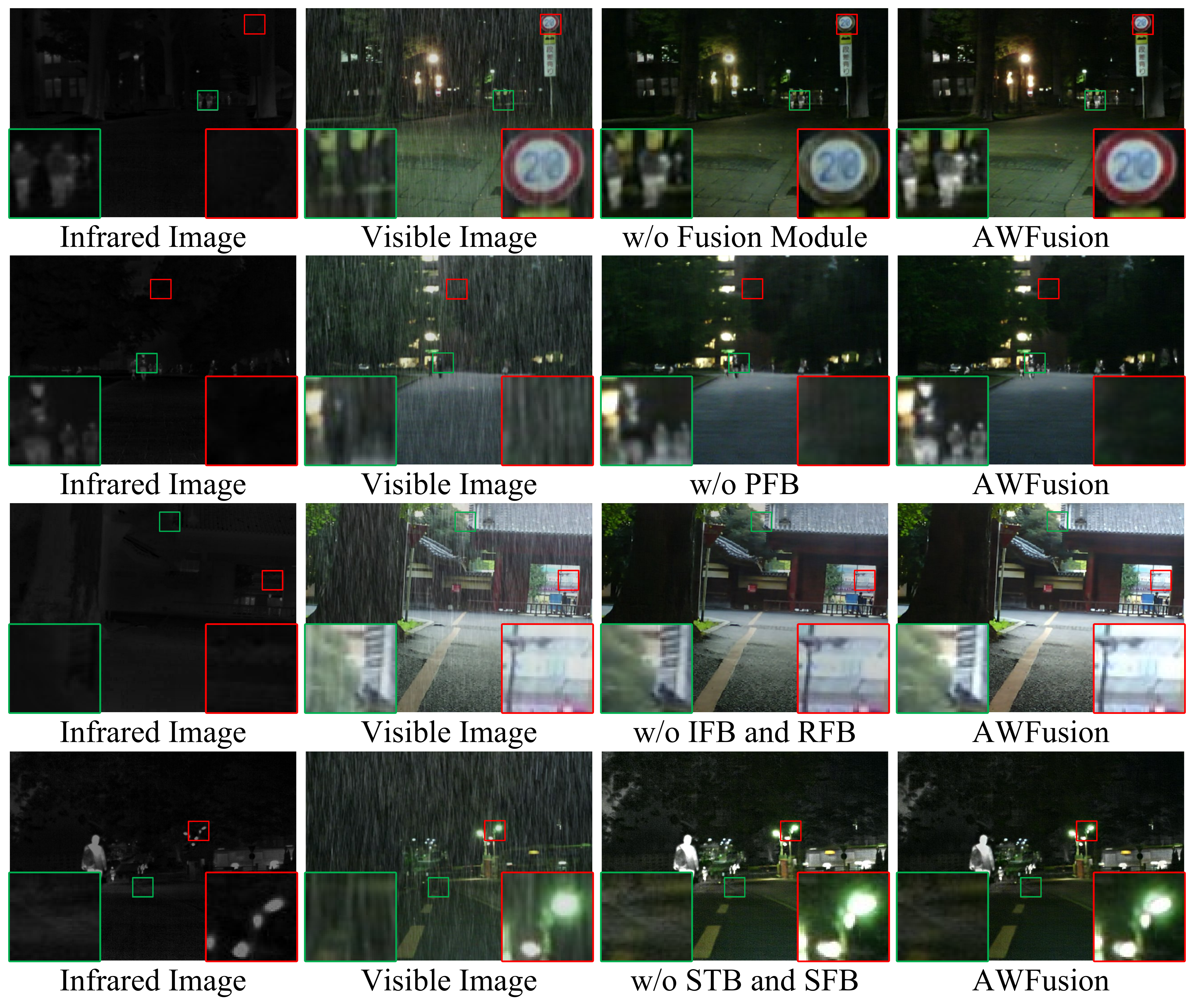}
   \caption{Qualitative comparison of ablation experiments under rain weather.}
   \label{fig12}
\end{figure}

\begin{table*}[h]
\caption{Quantitative comparison results of different methods under multiple weather disturbances. \textbf{Bold} is the best. }
\begin{adjustbox}{width=\textwidth}
\begin{tabular}{cccccccccccc}
\hline
\multicolumn{1}{c|}{Methods}       & \multicolumn{1}{c|}{Pub.}             & \multicolumn{1}{c|}{Restor.}                & $Q_{G}$↑                           & $Q_{M}$↑                            & $Q_{S}$↑                            & $Q_{CB}$↑                           & $Q_{CV}$↓               &  $Q^{AB/F}$↑             & $SSIM$↑                          & $SCD$↑                                                    & $EN$↑              \\ \hline
\multicolumn{12}{c}{Snow and Haze}                                                                                                                                                                                                                                                           \\ \hline
\multicolumn{1}{c|}{ControlFusion \cite{r101}} & \multicolumn{1}{c|}{\textit{NIPS 25}} & \multicolumn{1}{c|}{\multirow{3}{*}{MODEM \cite{r103}}} & 0.3408          & 0.3230          & 0.7097          & 0.4499          & \textbf{383.7061} & 0.3097          & 1.0466          & 0.4581          & 7.1471          \\
\multicolumn{1}{c|}{GIFNet \cite{r86}}        & \multicolumn{1}{c|}{\textit{CVPR 25}} & \multicolumn{1}{c|}{}                       & 0.2950          & 0.2811          & 0.6826          & 0.4579          & 717.2248          & 0.3304          & 1.2148          & 0.3595          & 6.6378          \\
\multicolumn{1}{c|}{SAGE \cite{r91}}          & \multicolumn{1}{c|}{\textit{CVPR 25}} & \multicolumn{1}{c|}{}                       & 0.3212          & \textbf{0.3630} & 0.7247          & 0.4445          & 443.1403          & 0.3259          & 1.1365          & 0.4524          & 7.0594          \\ \hline
\multicolumn{1}{c|}{AWFusion-S}    & \multicolumn{1}{c|}{\_}               & \multicolumn{1}{c|}{\_}                     & \textbf{0.3578} & 0.3590          & \textbf{0.7494} & \textbf{0.4584} & 519.5943          & \textbf{0.3698} & \textbf{1.3985} & \textbf{0.4656} & \textbf{7.1554} \\ \hline
\multicolumn{12}{c}{Rain and Haze}                                                                                                                                                                                                                                                           \\ \hline
\multicolumn{1}{c|}{ControlFusion \cite{r101}} & \multicolumn{1}{c|}{\textit{NIPS 25}} & \multicolumn{1}{c|}{\multirow{3}{*}{MODEM \cite{r103}}} & 0.3385          & 0.3453          & 0.7036          & 0.4350          & \textbf{385.4859} & 0.2995          & 0.9922          & 0.4680          & \textbf{7.3067} \\
\multicolumn{1}{c|}{GIFNet \cite{r86}}        & \multicolumn{1}{c|}{\textit{CVPR 25}} & \multicolumn{1}{c|}{}                       & 0.3023          & 0.2956          & 0.6924          & \textbf{0.4529} & 662.5418          & 0.3292          & 1.2082          & 0.3685          & 6.6517          \\
\multicolumn{1}{c|}{SAGE \cite{r91}}          & \multicolumn{1}{c|}{\textit{CVPR 25}} & \multicolumn{1}{c|}{}                       & 0.3182          & 0.3880          & 0.7221          & 0.4296          & 436.8800          & 0.3138          & 1.0881          & 0.4516          & 7.0836          \\ \hline
\multicolumn{1}{c|}{AWFusion-S}    & \multicolumn{1}{c|}{\_}               & \multicolumn{1}{c|}{\_}                     & \textbf{0.3775} & \textbf{0.3999} & \textbf{0.7530} & 0.4454          & 549.1318          & \textbf{0.3722} & \textbf{1.4212} & \textbf{0.4919} & 7.1886          \\ \hline
\multicolumn{12}{c}{Rain and Snow}                                                                                                                                                                                                                                                           \\ \hline
\multicolumn{1}{c|}{ControlFusion \cite{r101}} & \multicolumn{1}{c|}{\textit{NIPS 25}} & \multicolumn{1}{c|}{\multirow{3}{*}{MODEM \cite{r103}}} & \textbf{0.3551} & 0.3203          & 0.7258          & 0.4594          & 537.3284          & 0.3155          & 1.2346          & 0.4869          & 7.0407          \\
\multicolumn{1}{c|}{GIFNet \cite{r86}}        & \multicolumn{1}{c|}{\textit{CVPR 25}} & \multicolumn{1}{c|}{}                       & 0.2924          & 0.2826          & 0.6588          & 0.4622          & 1121.0535         & 0.3113          & 1.1707          & 0.3739          & 6.6209          \\
\multicolumn{1}{c|}{SAGE \cite{r91}}          & \multicolumn{1}{c|}{\textit{CVPR 25}} & \multicolumn{1}{c|}{}                       & 0.3138          & \textbf{0.3781} & \textbf{0.7579} & 0.4419          & 628.5957          & 0.3124          & 1.2184          & 0.4821          & 6.7914          \\ \hline
\multicolumn{1}{c|}{AWFusion-S}    & \multicolumn{1}{c|}{\_}               & \multicolumn{1}{c|}{\_}                     & 0.3382          & 0.3622          & 0.6955          & \textbf{0.4699} & \textbf{258.0324} & \textbf{0.3198} & \textbf{1.3459} & \textbf{0.4910} & \textbf{7.1263} \\ \hline
\end{tabular}
\label{tab_multi}
\end{adjustbox}
\end{table*}

\subsubsection{Object Detection Performance Comparison}
In real-world applications, systems must adapt to varying weather and environmental conditions. Although our method is tailored for severe weather, it also delivers high-quality fusion results in normal conditions, such as clear, haze-free, and rain-free environments. This demonstrates not only its effectiveness in harsh weather but also its strong generalization ability in clean settings.
In \cref{tab3}, we evaluated the object detection accuracies of all methods on the M3FD dataset using YOLOv7 \cite{r9}. As shown in \cref{tab3}, the proposed algorithm achieves the highest detection accuracy across all five categories, with both mAP@0.5 and mAP@[0.5:0.95] also obtaining the top scores.
These results confirm that the proposed algorithm retains sufficient information and detail for accurate localization and identification in object detection. Moreover, it enhances detection accuracy by preserving and leveraging complementary information.

\subsection{Extension to Comparisons with ``Restoration + Fusion'' pipelines}
To further validate the effectiveness of our method, we extend our experiments to comparisons with ``Restoration + Fusion'' pipelines.    Specifically, we first apply image restoration algorithms to recover the degraded visible images, and then feed the restored visible images together with clean infrared images into SOTA IVIF methods. In contrast, our model directly takes degraded multi-modal images as input.
We use two restoration methods, AdaIR \cite{r102} and MODEM \cite{r103}, and compare our method with five advanced IVIF algorithms: Text-DiFuse \cite{r96}, ControlFusion \cite{r101}, GIFNet \cite{r86}, SAGE \cite{r91}, and OmniFuse \cite{r98}. Qualitative results are shown in \cref{fig_restoration}, while quantitative results are reported in \cref{lab_AdaIR} and \cref{label_MODEM}. For fairness, both restoration algorithms were retrained on the AWMM-100K dataset.

From \cref{fig_restoration}, we can observe that although restoration algorithms can remove most degradation, the remaining artifacts and detail loss are inevitably passed to the subsequent fusion stage.  In rainy scenes, the fusion results of Text-DiFuse, ControlFusion, and OmniFuse show varying degrees of detail loss, leading to insufficient scene contrast.  GIFNet carries noticeable artifacts produced during restoration, which may interfere with downstream tasks such as target recognition and scene understanding.
In hazy scenes, similar issues persist.  GIFNet attempts to compensate for the lost details by overly enhancing infrared information, but this causes the final image to deviate from human visual perception.  Since Text-DiFuse and OmniFuse cannot perceive degradation cues, they amplify the residual degradation from the restoration stage, further reducing image quality.
In snowy scenes, snowflakes not only occlude structural details but also introduce effects similar to overexposure.  As a result, GIFNet and SAGE fail to preserve scene brightness, leading to incomplete depth information.  In contrast, the proposed method consistently delivers robust fusion performance across all three adverse conditions.  It preserves fine details, maintains color fidelity, and effectively removes degradation, outperforming all “Restoration + Fusion’’ pipelines.

\cref{lab_AdaIR} and \cref{label_MODEM} present the quantitative comparisons using AdaIR-restored and MODEM-restored visible images.  As reflected in the qualitative analysis, Text-DiFuse and OmniFuse obtain relatively low scores due to their significant loss of details.  The proposed method surpasses all competing approaches across the three weather conditions and achieves the highest scores on most evaluation metrics.
Overall, the proposed method outperforms the “Restoration + Fusion’’ pipelines in both qualitative and quantitative comparisons, further demonstrating its advantages for real-world applications.

\subsection{Extension to Multiple Weather Disturbances}
Compound degradation increases scene complexity, as rain, haze, and snow exhibit distinct physical characteristics: rain and snow cause localized occlusion, while haze reduces overall contrast. When combined, these effects create more challenging degradation. Although specific patterns can address individual disturbances, the model must exhibit strong adaptability and generalization to handle multiple disturbances simultaneously.
To evaluate the generalization ability of the proposed algorithm under severe weather conditions, we conducted extended experiments with composite degradation. In the comparison methods, the fusion operation is performed on pre-processed images. We adopt MODEM, a state-of-the-art adverse-weather removal model, to eliminate degradations in the visible images before fusion. Since MODEM is trained on AWMM-100K, it is capable of identifying and removing all three types of degradations. In contrast, our method directly generates fused results from the original degraded multi-modal inputs, without relying on any pre-processing.

As shown in \cref{fig11}, when two types of degradations appear simultaneously in the source images, the degradation removal capability of MODEM is significantly reduced. The residual degradation is carried over to the fusion result, leading to suboptimal outcomes. In contrast, the proposed method reconstructs high-fidelity scene details under complex weather conditions, preserving key features while avoiding issues like information loss.
\cref{tab_multi} presents the quantitative results of different methods under multiple weather degradations. As shown, the proposed method achieves the highest scores on six or more metrics across all three types of composite degradation.

\subsection{Ablation Experiment}
We performed an ablation analysis on the six  modules to verify their effectiveness. The modules were grouped into four experimental sets, with the results visualized in \cref{fig12} and quantitatively compared in \cref{tab4}.

\subsubsection{Effectiveness Analysis of the Fusion Module}
The image fusion module plays a critical role in the end-to-end framework by enabling multi-modality feature interaction. To evaluate its effectiveness, we performed an ablation analysis. As illustrated in the first row of \cref{fig12}, removing the fusion module disrupts this interaction, causing salient infrared features to be occluded by blurred visible details, as seen in the green zoomed-in area. Additionally, the absence of feature interaction introduces low-contrast IR information into the fusion output, making it difficult to reconstruct color information and resulting in slight color distortion, highlighted in the red zoomed-in area.

\begin{table}[t]
\caption{Quantitative comparison of ablation experiments in rain weather. \textbf{Bold} is the best.}
\resizebox{\columnwidth}{!}{%
\centering
\begin{tabular}{c|ccccc}
\hline
                  & {$Q_{M}$} & {$Q_{S}$} & {$Q_{CV}$} & {$SSIM$} & {$EN$} \\ \hline
w/o PFB           & 0.5439                    & 0.7917                    & 579.3986                   & 0.3048                      & 6.2609                    \\
w/o Fusion & 0.4989                    & 0.7953                    & 577.4851                   & 0.2165                      & 6.2930                    \\
w/o IFB and RFB   & 0.4646                    & 0.8096                    & 501.8484                   & 0.2923                      & 5.9288                    \\
w/o STB and SFB   & 0.5454                    & 0.7979                    & 536.2539                   & 0.3904                      & \textbf{6.7516}           \\
AWFusion        & \textbf{0.6855}           & \textbf{0.8444}           & \textbf{464.7867}          & \textbf{0.3993}             & 6.3160                    \\ \hline
\end{tabular}
}
\label{tab4}
\end{table}

\subsubsection{Effectiveness Analysis of the PFB}
The PFB module, built on the atmospheric scattering model, effectively mitigates weather-induced interference and reconstructs clear scene features. We performed an ablation analysis. As shown in the second row of \cref{fig11}, removing the PFB module leaves residual degradation in the fusion result. In texture-rich regions, such as leaves, the network struggles to distinguish between degradation artifacts and actual textures, as highlighted in the red zoomed-in area, leading to detail loss and blurred edges.

\subsubsection{Effectiveness Analysis of the IFB and RFB}
To improve the ability of network to estimate light transmission, we designed the IFB and RFB modules based on retina theory and constrained feature generation by reconstructing the loss. As shown in the third row of \cref{fig11}, removing these modules results in reduced lighting in the fusion output, likely due to inaccurate transmission map estimation and insufficient luminance compensation. This makes the absence of light recovery more noticeable. Additionally, \cref{tab4} shows a significant drop in the $Q_{M}$, $SSIM$, and $EN$ metrics.

\subsubsection{Effectiveness Analysis of the Fusion Strategies}
To leverage the low-rank and sparse features decomposed by the LLRR block, we designed the SFB to fuse sparse components and the STB to integrate low-rank components through context extraction. An ablation study was performed by removing the SFB and STB, replacing them with a maximum-value fusion strategy. As more feature channels are extracted from the source image, the simple fusion strategy impedes effective cross-modal interaction, restricting the extraction of salient and complementary features. This causes pixel redundancy in the fused output, reflected in the excessively high $EN$ metric in \cref{tab4}. The enlarged region in \cref{fig11} further highlights the increased noise in the ground, which complicates object recognition.

\section{Discussion}

\noindent \textbf{Limitations and Future Work:}
The proposed algorithm offers an effective solution for IVIF under adverse weather conditions.  While it demonstrates strong fusion performance, our experiments reveal that its computational efficiency leaves room for further improvement.  Specifically, for an input of size 256×256, the algorithm requires 193 ms of inference time, with 889 GFLOPs and 19.359 million parameters.  In future work, we aim to optimize the computational efficiency of the model, thereby enhancing its applicability in real-time and resource-constrained scenarios.

\noindent \textbf{Conclusion:}
In this study, we propose an all-weather MMIF method, focusing on the relationship between multi-modality image fusion and image restoration. Our framework shifts the goal of the network from pixel-perfect recovery to maximizing the representation of key information from the scene. The end-to-end design comprises two main modules: a fusion module that decouples multi-modal features into low-rank and sparse components for interactive merging, and a restoration module guided by physical model to restore clear details from blur pixels, ensuring high-quality results.
Additionally, we introduce a large-scale benchmark dataset, AWMM-100k, tailored for multi-modality image processing under adverse weather conditions. Experimental results confirm that the AWFusion surpasses existing methods in both adverse and normal conditions, with outstanding performance in downstream tasks, demonstrating its strong practical value.

\section*{Acknowledgement}
This research was supported by the National Natural Science Foundation of China (No. 62201149), the Natural Science Foundation of Guangdong Province (No. 2024A1515011880), the Basic and Applied Basic Research of Guangdong Province (No. 2023A1515140077), the Research Fund of Guangdong-HongKong-Macao Joint Laboratory for Intelligent Micro-Nano Optoelectronic Technology (No. 2020B1212030010).

 \bibliographystyle{elsarticle-num} 
 \bibliography{main}

\end{document}